\documentclass{article}

\usepackage{arxiv}

\usepackage[utf8]{inputenc} % allow utf-8 input
\usepackage[T1]{fontenc}    % use 8-bit T1 fonts
\usepackage{hyperref}       % hyperlinks
\usepackage{url}            % simple URL typesetting
\usepackage{booktabs}       % professional-quality tables
\usepackage{amsmath,amssymb,amsfonts}       % blackboard math symbols
\usepackage{nicefrac}       % compact symbols for 1/2, etc.
\usepackage{microtype}      % microtypography
\usepackage{lipsum}		% Can be removed after putting your text content
\usepackage{graphicx}
\usepackage[numbers]{natbib}
\usepackage{doi}
\usepackage{subfigure}  %插入多图时用子图显示的宏包
\usepackage[all]{hypcap}
\usepackage{mathrsfs}
\usepackage[capitalise]{cleveref}
\usepackage{float}%提供float浮动环境}
\usepackage{multirow}%提供跨行命令\multirow{}{}{}
\usepackage{makecell}% 三线表-竖线
\usepackage{algorithmic}

\title{MM-GTUNets: Unified Multi-Modal Graph Deep Learning for Brain Disorders Prediction\thanks{Accepted by IEEE Transactions on Medical Imaging.}}

\author{ \href{https://orcid.org/0009-0002-3150-6664}{\includegraphics[scale=0.06]{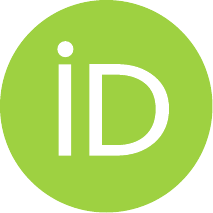}\hspace{1mm}Luhui Cai}\\
    Shanghai Maritime University\\
	Shanghai 201306, China \\
	\texttt{clh0x123@126.com} \\
	\And
	\href{https://orcid.org/0000-0002-9035-8078}{\includegraphics[scale=0.06]{orcid.pdf}\hspace{1mm}Weiming Zeng}\thanks{Corresponding author} \\
	Shanghai Maritime University\\
	Shanghai 201306, China \\
	\texttt{zengwm86@163.com} \\
    \And
	\href{https://orcid.org/0009-0004-2934-3135}{\includegraphics[scale=0.06]{orcid.pdf}\hspace{1mm}Hongyu Chen} \\
	Shanghai Maritime University\\
	Shanghai 201306, China \\
	\texttt{hongychen676@gmail.com} \\
    \And
	\href{https://orcid.org/0000-0002-9022-9049}{\includegraphics[scale=0.06]{orcid.pdf}\hspace{1mm}Hua Zhang} \\
	Shanghai Maritime University\\
	Shanghai 201306, China \\
	\texttt{zhanghua\_hhz@163.com} \\
     \And
	\href{https://orcid.org/0009-0008-5310-124X}{\includegraphics[scale=0.06]{orcid.pdf}\hspace{1mm}Yueyang Li} \\
	Shanghai Maritime University\\
	Shanghai 201306, China \\
	\texttt{lyy20010615@163.com} \\
     \And
	\href{https://orcid.org/0009-0001-4762-3390}{\includegraphics[scale=0.06]{orcid.pdf}\hspace{1mm}Yu Feng} \\
	Shanghai Maritime University\\
	Shanghai 201306, China \\
	\texttt{fyuchn@163.com} \\
     \And
	\href{https://orcid.org/0009-0000-2553-2183}{\includegraphics[scale=0.06]{orcid.pdf}\hspace{1mm}Hongjie Yan} \\
	Affiliated Lianyungang Hospital of Xuzhou Medical University\\
	Lianyungang 222002, China \\
	\texttt{yanhjns@gmail.com} \\
     \And
	\href{https://orcid.org/0000-0001-9991-5865}{\includegraphics[scale=0.06]{orcid.pdf}\hspace{1mm}Lingbin Bian} \\
	ShanghaiTech University\\
	Shanghai 201210, China \\
	\texttt{bianlb@shanghaitech.edu.cn} \\
     \And
	\href{https://orcid.org/0000-0002-2154-5996}{\includegraphics[scale=0.06]{orcid.pdf}\hspace{1mm}Wai Ting Siok} \\
	The Hong Kong Polytechnic University\\
	Hong Kong, SAR, China \\
	\texttt{wai-ting.siok@polyu.edu.hk} \\
     \And
	\href{https://orcid.org/0000-0002-9701-2918}{\includegraphics[scale=0.06]{orcid.pdf}\hspace{1mm}Nizhuan Wang}\footnotemark[2]\\
	The Hong Kong Polytechnic University\\
	Hong Kong, SAR, China \\
	\texttt{wangnizhuan1120@gmail.com} \\
}

\date{}

\renewcommand{\headeright}{}
\renewcommand{\undertitle}{}
\renewcommand{\shorttitle}{CAI\;\textit{et al.}: MM-GTUNets: Unified Multi-Modal Graph Deep Learning for Brain Disorders Prediction}

\Crefformat{figure}{#2Fig.~#1#3}%figure缩写
\Crefmultiformat{figure}{Figs.~#2#1#3}{ and~#2#1#3}{, #2#1#3}{ and~#2#1#3}
\Crefformat{section}{#2Sec.~#1#3}%section缩写
\Crefmultiformat{section}{Secs.~#2#1#3}{ and~#2#1#3}{, #2#1#3}{ and~#2#1#3}
\Crefformat{equation}{#2Eq.~#1#3}%equation缩写
\Crefmultiformat{equation}{Eqs.~#2#1#3}{ and~#2#1#3}{, #2#1#3}{ and~#2#1#3}

\newcommand{\customunderline}[2][0pt]{%下划线间距
  \raisebox{#1}{\underline{\smash{#2}}}%
}

\begin{document}
\maketitle

\begin{abstract}
	Graph deep learning (GDL) has demonstrated impressive performance in predicting population-based brain disorders (BDs) through the integration of both imaging and non-imaging data. However, the effectiveness of GDL-based methods heavily depends on the quality of modeling  multi-modal population graphs and tends to degrade as the graph scale increases. Moreover, these methods often limit interactions between imaging and non-imaging data to node-edge interactions within the graph, overlooking complex inter-modal correlations and resulting in suboptimal outcomes. To address these challenges, we propose MM-GTUNets, an end-to-end Graph Transformer-based multi-modal graph deep learning (MMGDL) framework designed for large-scale brain disorders prediction. To effectively utilize rich multi-modal disease-related information, we introduce \underline{M}odality \underline{R}eward \underline{R}epresentation \underline{L}earning (MRRL), which dynamically constructs population graphs using an Affinity Metric Reward System
    (AMRS). We also employ a variational autoencoder to reconstruct latent representations of non-imaging features aligned with imaging features. Based on this, we introduce \underline{A}daptive \underline{C}ross-\underline{M}odal \underline{G}raph \underline{L}earning (ACMGL), which captures critical modality-specific and modality-shared features through a unified GTUNet encoder, taking advantages of Graph UNet and Graph Transformer, along with a feature fusion module. We validated our method on two public multi-modal datasets ABIDE and ADHD-200, demonstrating its superior performance in diagnosing BDs. Our code will be available at \href{https://github.com/NZWANG/MM-GTUNets}{https://github.com/NZWANG/MM-GTUNets}.
\end{abstract}

% keywords can be removed
\keywords{Graph deep learning \and reward system \and cross-modal learning \and disease prediction}

\section{Introduction}
\label{sec:1}
Brain disorders (BDs), including neurodevelopmental conditions such as Autism Spectrum Disorder (ASD) and Attention Deficit Hyperactivity Disorder (ADHD) \cite{hirota2023autism, posner2020attentiondeficit}, often involve complex pathological mechanisms and diverse clinical manifestations. These disorders severely impact the quality of life and social functioning of subjects struggling with them. By 2021, over 3 billion subjects globally had been affected by different BDs, posing substantial challenges to the worldwide healthcare systems \cite{steinmetz2024global}. 

Currently, many clinical diagnosis of BDs are based on multi-modal medical data, but the diverse nature of the data poses a challenge for healthcare professionals in delivering accurate, reliable and timely diagnoses \cite{yanase2019systematic}. This challenge has been tackled through the development of novel Artificial Intelligence (AI) prediction models capable of processing large-scale multi-modal medical data for the clinical diagnosis of BDs \cite{acosta2022multimodal,salvi2024multimodality,saab2024capabilities}. For instance, \cite{huang2018crossmodality,wu2018sparse} used sparse dictionary learning to obtain the complementary features across modalities. The research conducted in \cite{yang2023learning,zhu2023deep} introduced specialized deep encoders designed for different modalities to acquire shared representations. However, these methods primarily focused on the complementarity and consistency across multi-modal data \cite{zhang2020multimodal}, neglecting the crucial relationships among subjects necessary for diagnosing BDs \cite{parisot2018disease,huang2022disease,kazi2019inceptiongcn,cosmo2020latentgraph}. 

Graph Deep Learning (GDL) \cite{cosmo2020latentgraph,kazi2022differentiable} introduces a new approach for predicting population-based BDs by integrating multi-modal information and elucidating connections among subjects. In this framework, subjects are nodes in a graph and the relationships between them are depicted as edges. These graph-based models update node features by aggregating information from the neighboring nodes  \cite{kipf2017semisupervised,gao2021graph}. The strategies for constructing population graphs can be either static or adaptive \cite{muller2024a}. \cite{parisot2018disease,kazi2019inceptiongcn} employed predefined similarity measures to evaluate correlations among subjects' multi-modal features. These methods rely on fixed similarity measures, limiting their adaptability during training and leading to suboptimal generalization performance. To address this limitation, \cite{huang2022disease,zhang2023classification} devised a pairwise association encoder to dynamically adjust the edge weights of the population graph during training.   \cite{song2023multicenter} examined the impact of various categories of non-imaging features on edge characteristics. They introduced an attention mechanism to compute affinity scores for these features and assign corresponding attention weights to enhance the capture of relational information between features.

Among various GDL strategies, the Graph Neural Network (GNN) emerges as a popular and effective method tailored for processing graph-structured data. GNNs can effectively capture and leverage the intricate relationships and dependencies within graph data and have been widely applied in population-based BDs prediction \cite{parisot2018disease,kazi2022differentiable,zhang2023classification}. Some researchers have suggested integrating Graph Transformer (GT) into GNNs to address the complexity of population graphs \cite{ying2021transformers,shi2021masked}. GT employs a self-attention mechanism that enables the model to focus on and dynamically weigh the nodes and edges, facilitating the learning and representation of inter-node relationships. This mechanism allows GNNs to extract information from the entire graph structure, not just local neighborhoods, fostering the integration of node and edge information and enabling efficient message passing within a graph. This innovation has prompted the exploration of various GT-based model architectures in BDs prediction research, demonstrating the continuous evolution and advancement in this domain \cite{pellegrini2023unsupervised,guan2024dynamic}.

Although the aforementioned methods have shown strong performance in population-based BDs prediction tasks, challenges persist due to the diversity and complexity of multi-modal data across different subjects.

\subsection{Underutilization of Non-Imaging Data}
\label{subsec:1.1}

Using both imaging data (e.g. the resting-state functional magnetic resonance imaging (rs-fMRI)) and non-imaging data (e.g., gender, age, acquisition site, etc.)  \cite{parisot2018disease,huang2022disease} can provide complementary information and helps in understanding BDs manifestations better. Yet, in many previous studies, non-imaging data were mainly used to calculate affinity scores (e.g., the connection strengths between nodes) of the population graph and were not included as part of the subject features when updating information in the GNN framework\cite{parisot2018disease,huang2022disease,kazi2019inceptiongcn}. This approach may underutilize crucial non-imaging data. To address this issue, previous studies\cite{cosmo2020latentgraph,zheng2022multimodal} have attempted to construct separate graphs for imaging and non-imaging data. However, due to significant differences between the two data types, simply incorporating low-dimensional non-imaging features as part of the graph nodes may not fully capture their latent information. Despite attempts like \cite{song2023multicenter} introducing an attention mechanism for preliminary exploration, the challenge remains that different non-imaging data types might influence the connection strengths in the group graph differently, posing an ongoing problem that needs resolution.

\subsection{Overlooking Crucial Node Features}
\label{subsec:1.2}
The performance of GNN is closely tied to the graph structure\cite{kipf2017semisupervised,wu2021comprehensive}. To enhance GNN effectiveness, it is crucial to reduce the impact of noise when constructing population graphs. Many existing graph contruction methods, both manual \cite{parisot2018disease,kazi2019inceptiongcn} and adaptive \cite{huang2022disease,zheng2022multimodal}, operate on large graphs with numerous subject nodes. However, these methods often overlook key node features and process the entire graph uniformly. A more effective approach involves integrating pooling layers to filter nodes and edges in the graph \cite{gao2021graph}. While this approach was used in a previous study on predicting BDs \cite{li2021braingnn}, its application in population-based BDs prediction remains unexplored.

\subsection{Insufficient Depth in Cross-Modal Interaction}
\label{subsec:1.3}
Cross-modal interaction refers to the process where information from one  modality (e.g., image, text, video, etc.) influences or interacts with another modality. In the diagnosis of BDs, imaging data reveals the changes of brain patterns, while non-imaging data reflects clinical and physiological characteristics. This interaction between modalities is vital for a thorough patient evaluation and effective diagnostic strategies. However, current approaches to cross-modal interaction between imaging and non-imaging data have limitations. Some methods like \cite{parisot2018disease, huang2022disease} focus on interactions within the graph's nodes and edges, while others, such as \cite{cosmo2020latentgraph,pellegrini2023unsupervised}, simply combime imaging and non-imaging features. These methods lack efficient integration and interaction of different data modalities, potentially leading to models that do not fully exploit the complementary and consistent information present in imaging and non-imaging data.

To address the aforementioned issues, we propose a multi-modal graph deep learning (MMGDL) framework for predicting BDs, termed Multi-Modal Graph TransUNets (MM-GTUNets).  This unified encoder-decoder-based MMGDL framework is effective in handling large-scale multi-modal data. The main contributions of this framework are as follows:

\begin{itemize}
\item \underline{M}odality-\underline{R}ewarding \underline{R}epresentation \underline{L}earning (MRRL): We propose MRRL to construct an adaptive reward population graph. This method generates latent representations of non-imaging features and precisely analyzes their contribution weights using a meticulously designed Affinity Metric Reward System (AMRS), thereby dynamically learning the population graph.
\item \underline{A}daptive \underline{C}ross-\underline{M}odal \underline{G}raph \underline{L}earning (ACMGL): ACMGL enables interactive learning between multi-modal data and captures complex inter- and intra-modal relationships. The Graph TransUNet (GTUNet) can filter important node features through node down-sampling and effectively extract both global and local information. This capability makes it well-suited for processing complex, large-scale graph data.
\item Visualization of Contribution Weights: We visualize the learned inter- and intra-modal contribution weights, which represent the relative importance of each modality. This visualization enables interpretable decision support in medical applications by helping clinicians and researchers understand how various modalities contribute to the final prediction.
\end{itemize}

\begin{figure}[!t]
\centering
\includegraphics[width=1\linewidth]{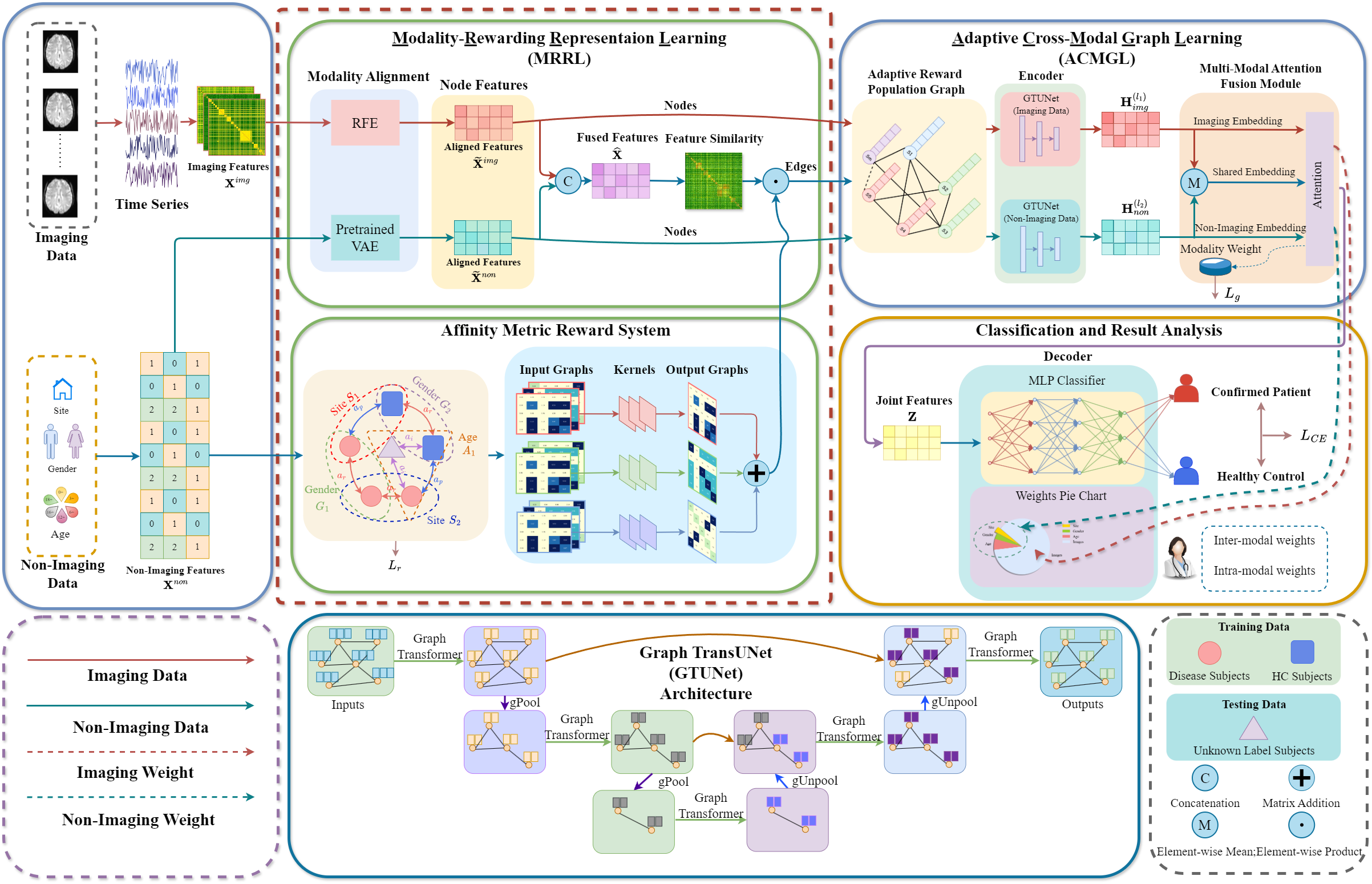}
\caption{The proposed MM-GTUNets framework for BDs prediction. In this paper, imaging data and non-imaging data refer to rs-fMRI and clinical data respectively. }
\label{fig:1}
\end{figure}

The remaining structure of the paper is outlined as follows: \Cref{sec:2} reviews related work on multi-modal graph deep learning. \Cref{sec:3} elaborates on the architecture of MM-GTUNets. \Cref{sec:4} describes the datasets and the implementation of results. \Cref{sec:5} discusses specific issues related to our proposed framework and outlines future work. Finally, a conclusion is presented in \Cref{sec:6}.

\section{RELATED WORK}
\label{sec:2}

\subsection{Graph Deep Learning}
\label{subsec:2.1}
Convolutional Neural Networks (CNNs) are capable of capturing multi-scale local features and constructing high-order representations \cite{lecun2015deep}. However, CNNs are mainly suitable for Euclidean data such as images and texts and are limited in dealing with non-Euclidean data such as social or brain networks \cite{wu2021comprehensive}. GNNs are better suited for processing graph-related data \cite{kipf2017semisupervised,ying2021transformers,shi2021masked}. In general, GNN-based methods for BDs diagnosis can be categorized into those based on brain graphs and those based on population graphs \cite{bessadok2023graph}.

\subsubsection{Brain Graphs}
\label{subsubsec:2.1.1}
The human brain network can be modeled as a graph, with each region of interest (ROI) serving as a node and the functional connectivity (FC) between ROIs representing the edge weight. For example, Brain-GNN uses an ROI-aware graph convolution layer and an ROI selection pooling layer to analyze brain networks for predicting neurobiomarkers at both group and subject levels \cite{li2021braingnn}. To enhance the efficiency of capturing brain connectome representations, \cite{zhao2022dynamic} introduced a dynamic graph convolution method and a convolutional pooling strategy for comprehensive graph information extraction, while \cite{zhang2023agcl} introduced graph contrastive learning for FC matrix feature extraction to identify key functional connections and brain regions associated with BDs.

\subsubsection{Population Graphs}
\label{subsubsec:2.1.2}
In population graphs, each subject is depicted as a node, with edge weights denoting the degree of association between subjects. This approach allows for the processing of each subject's feature representation on a single graph, enhancing efficiency and scalability compared to brain graph-based methods when handling large datasets \cite{parisot2018disease}. To address the challenge of label collection in clinical diagnosis, \cite{peng2023gate} introduced a self-supervised learning framework based on dynamic FC analysis. To tackle interpretability and biomarker detection challenges in population graph modeling, \cite{zhang2023classification,zaripova2023graph} proposed a hierarchical graph framework that accounts for both brain network topology and subject relationships.

\subsection{Multi-Modal Graph Deep Learning} 
\label{subsec:2.2}
Multi-modal graph deep learning (MMGDL) has been extensively applied in fields like computer vision and natural language processing \cite{ektefaie2023multimodal}. By integrating multi-modal data, MMGDL can decode more accurate and comprehensive information for predicting BDs. Based upon the work of \cite{parisot2018disease,huang2022disease}, \cite{zheng2022multimodal} proposed a modality-aware representation learning method for MMGDL. Similarly, \cite{chen2023ms} devised an MMGDL architecture incorporating a multi-modal fusion strategy. Both approaches aimed to capture inter- and intra-modal relationships. Research by  \cite{song2023multicenter,chen2024adversarial} delved into brain network connectomics analysis using multi-modal imaging data to unveil the structure and dynamic changes of brain functional networks, aiming at biomarker identification and the diagnosis of BDs.

\section{METHODOLOGY}
\label{sec:3}

\subsection{Problem Formulation}
\label{subsec:3.1}

\subsubsection{Overview of the Framework}
\label{subsubsec:3.1.1}
Our proposed end-to-end MM-GTUNets framework (\Cref{fig:1}) consists of three stages:
\begin{itemize}
    \item \textbf{\underline{M}odality-\underline{R}ewarding \underline{R}epresentation \underline{L}earning.} MRRL is designed to accurately construct the population graph's adjacency matrix $\mathbf{A}$ by aligning imaging features $\mathbf{X}^{img}$ and non-imaging features $\mathbf{X}^{non}$. Its reward metric system dynamically captures the significance of each type of non-imaging data.
    \item \textbf{\underline{A}daptive \underline{C}ross-\underline{M}odal \underline{G}raph \underline{L}earning.} Based on the modality-aligned features $\mathbf{\widetilde{X}}$ and the adjacency matrix $\mathbf{A}$, the unified encoder GTUNet and multi-modal attention module can achieve modality-joint representation $\mathbf{Z}$, consisting of both shared and specific modality information.
    \item \textbf{Classification Analysis.} This module uses a multi-layer perceptron (MLP) as an decoder to predict $\mathbf{\widehat{y}}$ from $\mathbf{Z}$. Also, the contribution weights of each modality in prediction are visualized.
\end{itemize} 

\subsubsection{Notation Definition}
\label{subsubsec:3.1.2}
Let $\mathbf{X}= [ \mathbf{X}_1, \mathbf{X}_2 \cdots, \mathbf{X}_N ] $ be the multi-modal features of $N$ subjects, and $\mathbf{y} = [ y_1, y_2, \cdots, y_N ]$ is the corresponding labels. For a subject $i$ with imaging and non-imaging features, we have $\mathbf{X}_i = [ \mathbf{x}_i^{img}, \mathbf{x}_i^{non}]$. Also, subject $i$ can be considered as node $v_i$, with its features represented as $\mathbf{X}_i$. Thus, the node set can be represented as $\mathcal{V} = \{ \langle{v_i}^{img}, v_i^{non}\rangle\}_{i=1}^N$. Further, the association strength between subject $i$ and subject $j$ is represented by the edge set $\mathcal{E} = \{ e_{ij} \}_{i,j=1}^{N}$. Therefore, the population graph is represented by $\mathcal{G} = (\mathcal{V}, \mathcal{E}, \mathbf{X})$, with its adjacency matrix $\mathbf{A} \in \mathbb{R}^{N \times N}$ showing the edge weights as $e_{ij}$.

\subsection{Modality-Rewarding Representation Learning}
\label{subsec:3.2}
\subsubsection{Modality Alignment}
\label{subsubsec:3.2.1}
To ensure the effectiveness of multi-modal features within the two channels of our proposed framework, we conducted comprehensive feature processing for both imaging and non-imaging data.

For the given imaging features $\mathbf{x}_{i}^{img}\in \mathbb{R}^{d_1}$ of subject $i$, we use the recursive feature elimination (RFE) dimensionality reduction strategy \cite{parisot2018disease} to convert into a relatively low-dimensional feature vector $\widetilde{\mathbf{x}}_{i}^{img}\in {{\mathbb{R}}^{d}}$. 

The low-dimensional non-imaging features $\mathbf{x}_{i}^{non}\in \mathbb{R}^{d_2}$ of subject $i$ exhibit a modal gap with the high-dimensional imaging features. To bridge the modal gap between imaging and non-imaging data, inspired by \cite{cohen2023joint} on multi-modal data imputation, we employed a pre-trained variational autoencoder (VAE)\cite{kingma2022autoencoding} to reconstruct the latent representation of $\mathbf{x}_{i}^{non}$, yielding a relatively high-dimensional vector $\widetilde{\mathbf{x}}_{i}^{non}\in {{\mathbb{R}}^{d}}$.

\subsubsection{Affinity Metric Reward System}
\label{subsubsec:3.2.2}
\begin{figure}
\centering
\includegraphics[width=0.6\linewidth]{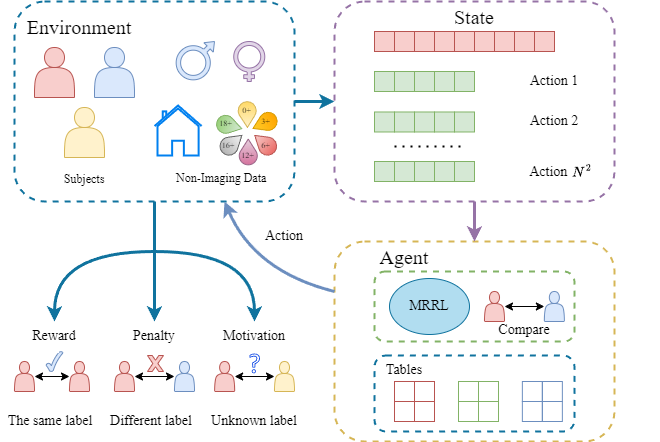}
\caption{Affinity Metric Reward System. AMRS adaptively adjusts the contribution weights of each type of non-imaging data and generates the non-imaging affinity graph, making the overall framework's diagnostic process more intelligent.}
\label{fig:2}
\end{figure}

Q-Learning is a classic reinforcement learning approach that enables an agent to learn the optimal policy by interacting with the environment, adjusting actions and updating the Q-table \cite{watkins1992qlearning}. It can be used to model the contribution weight ratios of various types of non-imaging data in AMRS (\Cref{fig:2}). For each pairwise comparison of subjects, their non-imaging information (state) and labels (action) are transmitted to MRRL (agent). Then, the agent selects the optimal operation according to the value, allowing specific non-imaging data to have a greater weight. The value is calculated based on the maintained three tables (i.e., reward, penalty, and motivation tables). By interacting with the subject population and their non-imaging data as the environment through the MRRL module, the AMRS learns the corresponding attention coefficients for each type of non-imaging data.

For convenience, we first give some mathematical notations in AMRS. The weights of the $v$ types of non-imaging data are defined as $\boldsymbol{\alpha}$:
\begin{equation}
    \begin{aligned}
        & \boldsymbol{\alpha }=\left[ {{\alpha }_{1}},{{\alpha }_{2}},\cdots ,{{\alpha }_{v}} \right], \\ 
        & \textit{s.t.}\left\{ 
        \begin{array}{*{35}{l}}
           0<{{\alpha }_{1}},{{\alpha }_{2}},\cdots ,{{\alpha }_{v}}<1,  \\
           \sum\limits_{u=1}^{v}{{{\alpha }_{u}}}=1.  \\
        \end{array} \right. \\ 
    \end{aligned}
    \label{eq:1}
\end{equation}

The reward, penalty, and motivation tables maintained by the agent are denoted as $\mathbf{R}, \mathbf{P}, \mathbf{M} \in {{\mathbb{R}}^{N\times N}}$, respectively. Also, the corresponding weight coefficients corresponding to $v$ types of non-imaging data are denoted as $\boldsymbol{\beta}$:
\begin{equation}
    \begin{aligned}
         & \boldsymbol{\beta} =\left\{ \beta _{r}^{u},\beta _{p}^{u},\beta _{m}^{u} \right\}_{u=1}^{v}, \\ 
         & \textit{s.t.}\left\{ 
         \begin{array}{*{35}{l}}
           \beta _{r}^{u},\beta _{m}^{u}>0,\beta _{p}^{u}<0,  \\
           \beta _{r}^{u}+\beta _{m}^{u}<\left| \beta _{p}^{u} \right|.  \\
        \end{array} \right. \\ 
    \end{aligned}
    \label{eq:2}
\end{equation}

Thus, the adjacency matrix of the non-imaging affinity graph can be defined as $\mathbf{C}\in {{\mathbb{R}}^{N\times N}}$, where ${{C}_{ij}}$ between subject $i$ and subject $j$ through AMRS is calculated as follow:
\begin{equation}
    {{C}_{ij}}=\textit{Sigmoid}\left( \sum\limits_{u=1}^{v}{{{\alpha }_{u}}}\left( \beta _{r}^{u}{{R}_{ij}}+\beta _{p}^{u}{{P}_{ij}}+\beta _{m}^{u}{{M}_{ij}} \right) \right).
    \label{eq:3}
\end{equation}
Meanwhile, the ${{R}_{ij}}$, ${{P}_{ij}}$ and ${{M}_{ij}}$ are defined as:
\begin{equation}
    \begin{aligned}
    &\Phi_{ij} = \sum_{u=1}^{v} \phi_{u}(u_{i}, u_{j}),\\
    \end{aligned}
\label{eq:4}
\end{equation}
where $(\Phi_{ij}, \phi_{u}) \in \{{ (R_{ij}, r_{u}), (P_{ij}, p_{u}), (M_{ij}, m_{u}) }\}$. 
Specifically, $\phi_{u}(u_{i}, u_{j})$ in \Cref{eq:4} can update the states of three tables after subjects $i$ and $j$ through interacting with the $u$-th non-imaging data according to \Cref{eq:5}. 
\begin{equation}
    \begin{aligned}
    r_{u}(u_{i}, u_{j}) &= \begin{cases}
        1, & \text{if } u_{i} = u_{j} \text{ and } y_{i} = y_{j}, \\
        0, & \text{otherwise},
    \end{cases} \\
    p_{u}(u_{i}, u_{j}) &= \begin{cases}
        1, & \text{if } u_{i} = u_{j} \text{ and } y_{i} \ne y_{j}, \\
        0, & \text{otherwise},
    \end{cases} \\
    m_{u}(u_{i}, u_{j}) &= \begin{cases}
        1, & \text{if } u_{i} = u_{j} \text{ and } \{ y_{i}, y_{j} \} \in \textit{testset}, \\
        0, & \text{otherwise}.
    \end{cases}
\end{aligned}
\label{eq:5}
\end{equation}
Finally, to adaptively determine the optimal weights $\boldsymbol{\alpha}$ in \Cref{eq:3} for each type of non-imaging data in constructing the affinity graph, a state-action value function $Q\left( s,a \right)$ is designed to maximize the overall value, which incorporates the policy $\pi$ as follow:
\begin{equation}
   \begin{aligned}
    Q_{\pi}(s, a) &= \mathbb{E}_{\pi} \left[ G_t \mid S_t = s, \mathcal{A}_t = a \right]  \\
    &= \frac{1}{N^2} \sum_{u=1}^{v} \underset{\alpha_{u}}{\mathop{\operatorname{argmax}}} \\ &\sum_{i=1}^{N} \sum_{j=1}^{N} \alpha_{u} \textit{ReLU} \left( \beta_{r}^{u} R_{ij} + \beta_{p}^{u} P_{ij} \right),
    \end{aligned}
\label{eq:6}
\end{equation}
where $G_t$, $S_t$, and $\mathcal{A}_t$ represent the cumulative returns, the weights of non-imaging features and the labels of the compared subjects from the beginning to the $t$-th comparison, respectively. Further, we can rewrite the optimization for the adjacency matrix $\mathbf{C}^*$ of the non-imaging affinity graph as: 
\begin{equation}
    {\mathbf{C}^{*}}\Leftrightarrow \underset{\pi }{\mathop{\operatorname{argmax}}}\,{{Q}_{\pi }}\left(s,a \right).
\label{eq:7}
\end{equation}

\subsubsection{Population Graph Construction}
\label{subsubsec:3.2.3}
Many studies have asserted that the fusion of cross-modal data can enhance the representation ability and improve the model's performance \cite{song2023multicenter, zheng2022multimodal, chen2023ms}. Given the imaging features $\widetilde{\mathbf{X}}^{img} \in \mathbb{R}^{N \times d}$, non-imaging features $\widetilde{\mathbf{X}}^{non} \in \mathbb{R}^{N \times d}$, and the adjacency matrix $\mathbf{C}$ of non-imaging affinity graph, the adaptive reward population graph (ARPG) construction is proposed to integrate key features of different modalities through the following steps:
\begin{itemize}
    \item Step 1: Forming the node embeddings of ARPG by directly fusing the features from various modalities through  $\mathbf{\widehat{X}}=\textit{Concat}({\widetilde{\mathbf{X}}}^{img},{\widetilde{\mathbf{X}}}^{non})$.
    \item Step 2: Generating the adjacency matrix $\mathbf{A}$ of ARPG. $\widehat{\mathbf{x}}_i$  denotes the fused feature vector of subject $i$. By integrating the non-imaging affinity graph $\mathbf{C}$ and the subject similarity measurement from \cite{parisot2018disease}, each edge weight of the ARPG is computed as:
    \begin{equation}
    {{A}_{ij}}=\textit{Sim}\left( {{{\widehat{\mathbf{x}}}}_{i}},{{{\widehat{\mathbf{x}}}}_{j}} \right)\odot {{C}_{ij}},
      \label{eq:8}
       \end{equation}
     where $\odot $ represents element-wise multiplication, and the similarity measurement $\textit{Sim}\left( \cdot  \right)$ is defined as follows:
    \begin{equation}
    \textit{Sim}\left( {{\widehat{\mathbf{x}}}_{i}},{{\widehat{\mathbf{x}}}_{j}} \right)=\textit{exp} \left( -\frac{{{\left[ \rho \left( {{\widehat{\mathbf{x}}}_{i}},{{\widehat{\mathbf{x}}}_{j}} \right) \right]}^{2}}}{2{{\sigma }^{2}}} \right),
    \label{eq:9}
   \end{equation}
    where $\rho(\cdot)$ represents the correlation distance function, and $\sigma$ denotes the width of the kernel. 
\end{itemize}
In addition, inspired by the methodology \cite{huang2022disease}, the Monte Carlo edge dropout strategy is integrated into our framework to randomly remove edges during training, inducing sparsity. This strategy can help mitigate over-smoothing and overfitting problems.

\subsection{Adaptive Cross-Modal Graph Learning}
\label{subsec:3.3}

\subsubsection{GTUNet Encoder}
\label{subsubsec:3.3.1}
\begin{figure}[!t]
\centering
\includegraphics[width=0.8\linewidth]{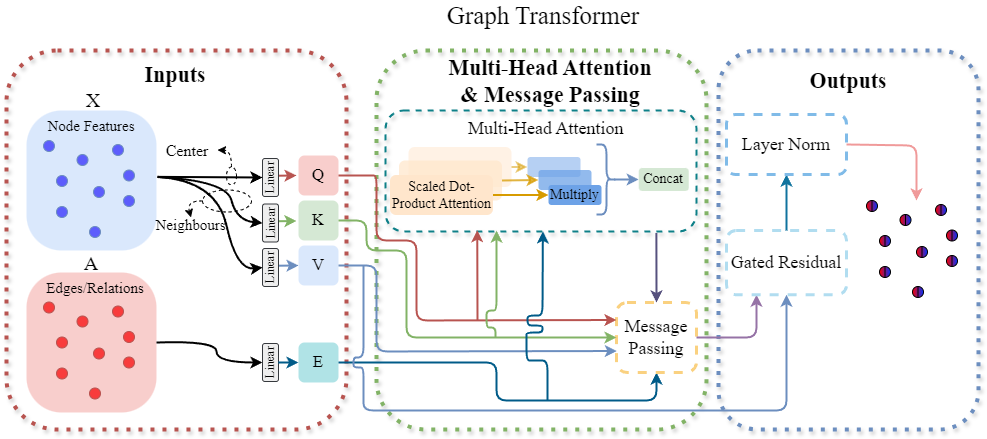}
\caption{Graph Transformer (GT)\cite{ying2021transformers,shi2021masked} architecture: The GT layer improves global context capture through self-attention, utilizes multi-head attention for diverse feature learning, and prevents over-smoothing with residual connections and layer normalization, enabling deeper architectures and enhanced performance.}
\label{fig:3}
\end{figure}

In ACMGL module, the GTUNet encoder is proposed to effectively extract modality-specific information from each modality channel, where the Graph Unets architecture \cite{gao2021graph} is used. The GTUNet includes the gPool layer for node downsampling, and the gUnpool layer for the corresponding inverse operation. The GT layer \cite{ying2021transformers,shi2021masked}, as depicted in \cref{fig:3} is used to replace the original graph convolution layer in Graph Unets. Details of the GTUNet architecture are illustrated in \cref{fig:1}.

Firstly, given the constructed ARPG, the updating process of node features in GTUNet can be denoted as an operation $\textit{GTC}\left( \cdot \right)$ through the GT layer, which can be concretely expressed as the following formulas. Let $\mathbf{H}^{(l)}=[\mathbf{h}_1^{(l)}, \mathbf{h}_2^{(l)},\cdots, \mathbf{h}_N^{(l)} ]$ represent the $l$-th layer of node embeddings, where $\mathbf{H}^{(0)}$ corresponds to the input features of the nodes $\widetilde{\mathbf{X}} = \left[ \widetilde{\mathbf{X}}^{img}, \widetilde{\mathbf{X}}^{non} \right]$ in ARPG. The attention for each edge from node $i$ to node $j$ can be computed as:
\begin{equation}
    \begin{aligned}
        \boldsymbol{\chi}_{i}^{(l)}\;\;\; &= \mathbf{W}_{\chi}^{(l)} \mathbf{h}_{i}^{(l)} + \mathbf{b}_{\chi}^{(l)}, \;\; \text{for} \;\; \boldsymbol{\chi} \in \{\mathbf{q}, \mathbf{k}, \mathbf{v}\}, \\
        \mathbf{e}_{ij}\;\;\;\; &= \mathbf{W}_{e} \mathbf{e}_{ij} + \mathbf{b}_{e}, \\
        \boldsymbol{\alpha}_{ij}^{(l)}\;\;\; &= \frac{\left\langle \mathbf{q}_{i}^{(l)}, \mathbf{k}_{j}^{(l)} + \mathbf{e}_{ij} \right\rangle}{\sum\limits_{u \in \mathcal{N}(i)} \left\langle \mathbf{q}_{i}^{(l)}, \mathbf{k}_{u}^{(l)} + \mathbf{e}_{iu} \right\rangle},\\
        \bar{\mathbf{h}}_{i}^{(l+1)} &= \sum_{j \in \mathcal{N}(i)} \boldsymbol{\alpha}_{ij}^{(l)} \left( \mathbf{v}_{j}^{(l)} + \mathbf{e}_{ij} \right),
    \end{aligned}
\label{eq:10}
\end{equation}
where $\mathcal{N}(i)$ denotes all the neighbors of node $i$, $\left\langle \mathbf{q},\mathbf{k} \right\rangle = \text{exp}\left( \frac{{{\mathbf{q}}^{\text{T}}}\mathbf{k}}{\sqrt{d}} \right)$ represents the scaled dot-product function, and $d$ is the dimension of the attention hidden layer. To avoid the over-smoothing of the model, the gated residual connections are introduced as follow: 
\begin{equation}
    \begin{aligned}
    \mathbf{r}_{i}^{(l)}\;\;\;\; &= \mathbf{W}_{r}^{(l)} \mathbf{h}_{i}^{(l)} + \mathbf{b}_{r}^{(l)}, \\
    \gamma_{i}^{(l)}\;\;\;\; &= \textit{Sigmoid}\left( \mathbf{W}_{g}^{(l)} \left[ \bar{\mathbf{h}}_{i}^{(l+1)}; \mathbf{r}_{i}^{(l)}; \bar{\mathbf{h}}_{i}^{(l+1)} - \mathbf{r}_{i}^{(l)} \right] \right), \\
    \mathbf{h}_{i}^{(l+1)} &= \textit{ReLU}\left( \textit{LN}\left( (1 - \gamma_{i}^{(l)}) \bar{\mathbf{h}}_{i}^{(l+1)} + \gamma_{i}^{(l)} \mathbf{r}_{i}^{(l)} \right) \right).
    \end{aligned}
\label{eq:11}
\end{equation}

Secondly, the downsampling and upsampling processes of GTUNet will be elaborated based on ARPG. Let ${\mathbf{A}}^{(l)}$ denote the $l$-th layer adjacency matrix, with ${\mathbf{A}}^{(0)}=\mathbf{A}$. In terms of the downsampling processing of GTUNet, the node features ${\mathbf{H}}^{(l)}$ and adjacency matrix ${\mathbf{A}}^{(l)}$ are first fed into the gPool layer \cite{gao2021graph}, followed by utilizing the top-k algorithm to select the $\widehat{k}$ nodes with the highest information, yielding the filtered node features ${\widehat{\mathbf{H}}}^{(l)}$ and adjacency matrix ${\widehat{\mathbf{A}}}^{(l)}$ as follows:
\begin{equation}
    \begin{aligned}
    \text{idx}\;\; &= \textit{rank}(\boldsymbol{\delta}, \widehat{k}), \\
    \widehat{\mathbf{A}}^{(l)} &= \mathbf{A}^{(l)}(\text{idx}, \text{idx}), \\
    \widehat{\mathbf{H}}^{(l)} &= \mathbf{H}^{(l)}(\text{idx}, :) \odot \left( \textit{Sigmoid}(\boldsymbol{\delta}(\text{idx}))\: \mathbf{1}_{d_{h}}^{T} \right),
    \end{aligned}
\label{eq:12}
\end{equation}
where $\boldsymbol{\delta}$ represents the projection of $\mathbf{H}^{(l)}$ on the learnable vector, $\widehat{k}$ represents the number of nodes selected in the new graph, $\textit{rank}(\boldsymbol{\delta}, \widehat{k})$ is a node ranking operation that returns the indices of the $\widehat{k}$ largest values in $\boldsymbol{\delta}$, and $\text{idx}$ denotes the indices selected in the new graph. $\mathbf{A}^{(l)}(\text{idx}, \text{idx})$ and $\mathbf{H}^{(l)}(\text{idx}, :)$ respectively represent the row and column extraction to form the feature matrix $\widehat{\mathbf{H}}^{(l)}$ and adjacency matrix $\widehat{\mathbf{A}}^{(l)}$ as inputs to $\textit{GTC}(\cdot)$. Notably, when $l$ is 0, $\widehat{\mathbf{H}}^{(0)} = \mathbf{H}^{(0)}$ and $\widehat{\mathbf{A}}^{(0)} = \mathbf{A}^{(0)}$. $\boldsymbol{\delta}(\text{idx})$ extracts the values of $\boldsymbol{\delta}$ with indices $\text{idx}$ and applies the $\textit{Sigmoid}(\cdot)$ operation. $\mathbf{1}_{d_h}^{T} \in \mathbb{R}^{d_h}$ is a one-dimensional tensor of size $d_h$ with all components having a value of 1, where $d_h$ is the neuron number at the hidden layer. $\odot$ denotes element-wise multiplication.

After the downsampling process that extracts important node features using $l$ gPool layers, we also need to use $l$ gUnpool layers to restore the graph to its original structure for subsequent classification tasks. Thus, the downsampling processing of GTUNet can be expressed by performing $\theta$ gUnpool layers:
\begin{equation}
    \widetilde{\mathbf{H}}^{(l+\theta)} = \textit{Distribute}\left( \mathbf{H}^{(l-\theta)}, \mathbf{H}^{(l+\theta)}, \text{idx}^{(l-\theta)} \right),
\label{eq:13}
\end{equation}
where $\widetilde{\mathbf{H}}^{(l+\theta)}$ represents the restored feature matrix after executing the gUnpool layer at the $(l+\theta)$-th layer. Due to the symmetry of GTUNet structure, the pre-filtered feature matrix and index are denoted as $\mathbf{H}^{(l-\theta)}$ and $\text{idx}^{(l-\theta)}$, with the corresponding adjacency matrix denoted as $\mathbf{A}^{(l-\theta)}$. The $\textit{Distribute}(\cdot)$ operation distributes the row vectors from $\mathbf{H}^{(l+\theta)}$ to the feature matrix $\mathbf{H}^{(l-\theta)}$ according to the indices stored in $\text{idx}^{(l-\theta)}$.

Finally, after obtaining the feature matrix $\mathbf{H}^{(2l)}$ restored to its initial structure through $l$ gUnpool layers, we perform addtional $\textit{GTC}(\cdot)$ operation. In GTUNet, a total of $2l+1$ $\textit{GTC}(\cdot)$ operations are performed, resulting in $2l+1$ GT layers. For simplicity, we denote the overall process of GTUNet encoder as an operation $\textit{GTU}(\cdot)$.

\subsubsection{Multi-Modal Attention Fusion Module}
\label{subsubsec:3.3.2}
Inspired by \cite{chen2023ms} and based on the outputs by the GTUNet encoder for each modality, we propose a multi-modal attention fusion module to effectively achieve modality-joint representations that encompass both modality-specific and shared information. This can be formulated as follows:
\begin{equation}
    \begin{aligned}
        \mathbf{Z}_{sp}^{img}  &=  \textit{GTU}^{(l_1)}(\widetilde{\mathbf{X}}^{img}) ,  \\   
        \mathbf{Z}_{sp}^{non}  &= \textit{GTU}^{(l_2)}(\widetilde{\mathbf{X}}^{non}),  \\   
        \mathbf{Z}_{sh}\;\;  &= \frac{1}{2} (\mathbf{Z}_{sp}^{img} + \mathbf{Z}_{sp}^{non}),
    \end{aligned}
\label{eq:14}
\end{equation}
where $l$ represents the layers of the encoder, $\mathbf{Z}_{sp}^{img}$ and $\mathbf{Z}_{sp}^{non}$ are the modality-specific embeddings for the imaging and non-imaging features, respectively, and $\mathbf{Z}_{sh}$ is the modality-shared embedding.

Next, the attention weights of the modality-shared embedding and two modality-specific embeddings can be expressed as:
\begin{equation}
    \begin{aligned}
   \boldsymbol{\tau }_{sh}\;\;  &= \textit{tanh} \left( \mathbf{W}{{\mathbf{Z}}_{sh}}+\mathbf{B} \right),  \\
   \boldsymbol{\tau }_{sp}^{img}  &= \textit{tanh} \left( {{\mathbf{W}}^{img}}\mathbf{Z}_{sp}^{img}+{{\mathbf{B}}^{img}} \right),  \\
   \boldsymbol{\tau }_{sp}^{non}  &= \textit{tanh} \left( {{\mathbf{W}}^{non}}\mathbf{Z}_{sp}^{non}+{{\mathbf{B}}^{non}} \right),  \\
    \end{aligned}
\label{eq:15}
\end{equation}
where $\mathbf{W}$, $\mathbf{W}^{img}$, and ${{\mathbf{W}}^{non}}$ represent the weight matrices, while $\mathbf{B}$, $\mathbf{B}^{img}$, and ${{\mathbf{B}}^{non}}$ denote the bias matrices.

After obtaining the attention weights, i.e., ${{\mathbf{Z}}_{sh}}$, $\mathbf{Z}_{sp}^{img}$ and $\mathbf{Z}_{sp}^{non}$, the final embedding can be derived by combining these weights with the embeddings as below:
\begin{equation}
    \begin{aligned}
   \mathbf{Z} &= {{\boldsymbol{\tau }}_{sh}} \odot {{\mathbf{Z}}_{sh}} 
   + \boldsymbol{\tau }_{sp}^{img} \odot \mathbf{Z}_{sp}^{img} 
   + \boldsymbol{\tau }_{sp}^{non} \odot \mathbf{Z}_{sp}^{non},
    \end{aligned}
\label{eq:16}
\end{equation}
where $\odot$ signifies the element-wise multiplication, and $\mathbf{Z}$ represents the joint representation of multi-modal features. 

\subsection{Classification and Result Analysis}
\label{subsec:3.4}
For classification stage, the final prediction process can be achieved as $\mathbf{\widehat{y}}=\textit{MLP}(\mathbf{Z})$ based on the joint modality representation $\mathbf{Z}$.

Meanwhile, we compute the contribution weight scores for each modality based on the attention weights in \Cref{eq:15} as:
\begin{equation}
   \begin{aligned}
      \begin{aligned}
      {\boldsymbol{\omega}}\;\; =& ({\omega}_{img}, {\omega}_{non})\\
      =& \textit{Softmax}\left ( \textit{f}\left(\boldsymbol{\tau }_{sp}^{img}, \boldsymbol{\tau }_{sh}\right), \textit{f}\left(\boldsymbol{\tau }_{sp}^{non}, \boldsymbol{\tau }_{sh}\right) \right), \\
    {\textit{f}\left(\cdot\right)} =& \frac{\textit{tr}( \boldsymbol{\tau }_{sp}, \boldsymbol{\tau }_{sp})}
       {\textit{tr}( \boldsymbol{\tau }_{sh}, \boldsymbol{\tau }_{sh})}, 
  \end{aligned}
    \end{aligned}
\label{eq:17}
\end{equation}
where ${\omega}_{img}$ and ${\omega}_{non}$ represent the weights for imaging features and non-imaging features, respectively.

\subsection{Objective Function}
\label{subsec:3.5}
The objective function for MM-GTUNets is as follows:
\begin{equation}
    \begin{aligned}
   {\mathcal{L}_{total}} =& {\mathcal{L}_{ce}} 
   + {{\omega}_{img}} \mathcal{L}_{g}^{img} 
   + {{\omega }_{non}} \left( \mathcal{L}_{g}^{non} + \eta {\mathcal{L}_{r}} \right),
    \end{aligned}
\label{eq:18} 
\end{equation}
where ${\mathcal{L}_{ce}}$, $\mathcal{L}_{g}^{img}$, and $\mathcal{L}_{g}^{non}$ denotes the cross-entropy loss, and the graph regularization terms for imaging and non-imaging features, respectively;  ${\omega_{img}}$, ${\omega_{non}}$ are the learnable parameters, and $\eta $ stands for the hyperparameter.

\subsubsection{Graph Regularization}
\label{subsubsec:3.5.1}
The structure of the graph significantly impacts the performance of GDL. In adaptive graph learning, fine-tuning the smoothness and sparsity emerges as a pivotal concern. Inspired by \cite{zheng2022multimodal}, we design the graph regularization as follows:
\begin{equation}
    \begin{aligned}
   {\mathcal{L}_{g}^{\psi}}\;\; &= \lambda \mathcal{L}_{smh}^{\psi} + \mu {\mathcal{L}_{deg}}, \\
   {\mathcal{L}_{\text{smh}}^{\psi}} &= \frac{1}{2{{N}^{2}}} \sum\limits_{i,j=1}^{N} {{A}_{ij}} \left\| {{\mathbf{z}}_{i}^{\psi}} - {{\mathbf{z}}_{j}^{\psi}} \right\|_{2}^{2}, \\
   {\mathcal{L}_{\text{deg}}} &= -\frac{1}{N} {{\mathbf{1}}^{T}} \log (\mathbf{A} \cdot \mathbf{1}), \\
    \end{aligned}
\label{eq:19}
\end{equation}
where ${\psi} \in \{img, non\}$, ${\mathcal{L}_{smh}^{\psi}}$, and ${\mathcal{L}_{deg}}$ represent the smoothness regularization and the degree regularization of the graph along with the hyperparameters $\lambda $ and $\mu$. 

\subsubsection{Reward Regularization}
\label{subsubsec:3.5.2}
In \Cref{subsec:3.2}, we optimized the non-imaging affinity graph $\mathbf{C}$ using the designed state-action value function ${{Q}_{\pi }}(s,a)$. According to \Cref{eq:7}, optimizing $\mathbf{C}$ is equivalent to maximizing ${{Q}_{\pi }}(s,a)$ via policy $\pi$. Thus, we can derive:
\begin{equation}
    \begin{aligned}
       \operatorname{argmin}{\mathcal{L}_{r}} &= \operatorname{argmax}{{Q}_{\pi }}(s,a), \\
       &= \operatorname{argmin}\left( \frac{1}{{{Q}_{\pi }}(s,a)} \right). \\
    \end{aligned}
\label{eq:20}
\end{equation}

\section{EXPERIMENTAL RESULTS AND ANALYSIS}
\label{sec:4}

\subsection{Datasets and Pre-Processing}
\label{subsec:4.1}
Our proposed MM-GTUNets has been evaluated on two public brain imaging datasets: ABIDE and ADHD-200. The corresponding demographic information is detailed in \Cref{tab:1}. In this study, we utilized rs-fMRI scans as the neuroimaging data. Due to a high proportion of fields being missing or invalid in the original data, we only include the subjects' gender, age and acquisition site as non-imaging data.

\subsubsection{ABIDE}
\label{subsubsec:4.1.1}
The Autism Brain Imaging Data Exchange (ABIDE) community \cite{dimartino2014autism} has compiled neuroimaging and non-imaging data from 20 different international acquisition sites. To ensure a fair comparison with previous state-of-the-art (SOTA) methods \cite{parisot2018disease,huang2022disease, zheng2022multimodal}, we selected a cohort of 871 subjects from ABIDE, comprising 468 healthy controls (HC) and 403 ASD patients.

\subsubsection{ADHD-200}
\label{subsubsec:4.1.2}
The ADHD-200 dataset \cite{BELLEC2017275} includes rs-fMRI data and corresponding non-imaging data from 8 international acquisition sites. Following the methodology of \cite{zhao2022dynamic}, we chose data from four sites: New York University Medical Center, Peking University, Kennedy Krieger Institute, and the University of Pittsburgh. Due to  missing data in some samples, we ultimately included 582 subjects consisting of 364 HC and 218 ADHD patients.

\subsubsection{Pre-Processing}
\label{subsubsec:4.1.3}
The data pre-processing procedure mainly followed the methodology of Parisot et al.\cite{parisot2018disease}. For the rs-fMRI data from ABIDE and ADHD-200, preprocessing was conducted according to the C-PAC\footnote{\href{https://github.com/preprocessed-connectomes-project/abide}
{https://github.com/preprocessed-connectomes-project/abide}}  \cite{cameron2013automated} and Athena\footnote{\href{https://www.nitrc.org/plugins/mwiki/index.php/neurobureau:AthenaPipeline}{https://www.nitrc.org/plugins/mwiki/index.php/neurobureau:AthenaPipeline}} \cite{BELLEC2017275} pipeline configurations, respectively. Using the Anatomical Automatic Labeling (AAL) brain atlas \cite{rolls2020automated}, we segmented the rs-fMRI data into 116 ROIs. For each ROI, we computed the average time series and then calculated the FC matrix for each subject using Pearson's correlation coefficient. Finally, we extracted the upper triangular elements of each FC matrix and flattened them into a one-dimensional vector $\mathbf{x}^{img}$.

For the corresponding non-imaging data, the categorical data  were encoded in ordinal form, while the numerical data were directly converted into float values. Then, these data were concatenated into a one-dimensional vector $\mathbf{x}^{non}$.

\begin{table}[ht]
\renewcommand{\arraystretch}{1.15}
\centering
\caption{Demographic Statistics of the Datasets Used in This Work}
\begin{tabular}{c c c c c}
    \toprule
    \textbf{Dataset} & \textbf{Diagnosis} & \textbf{Subject} & \textbf{Gender} & \textbf{Age} \\ 
    & & & (Female / Male) & (Mean $\pm$ Std.)  \\ \midrule
    \multirow{2}{*}{\text{ABIDE}} & HC & 468 & 90 / 378 & 16.84 $\pm$ 7.23 \\  
     & ASD & 403 & 54 / 349 & 17.07 $\pm$ 7.95 \\ \midrule
    \multirow{2}{*}{\text{ADHD-200}} & HC & 364 & 166 / 198 & 12.42 $\pm$ 8.62 \\  
     & ADHD & 218 & 39 / 179 & 11.56 $\pm$ 5.91 \\ 
    \bottomrule
\end{tabular}
\label{tab:1}
\end{table}

\subsection{Implementation Details}
\label{subsec:4.2}
Our MM-GTUNets runs on a server equipped with 12 NVIDIA GeForce 4090 GPUs and is deployed on the PyTorch with Adam optimizer \cite{kingma2017adam}, with a total of 1.19M trainable parameters. For the initial pre-training of the VAE, we set the learning rate to be 1e-3, the weight decay rate to be 5e-4, and the model is trained for 3000 epochs. After freezing the pre-trained VAE, we optimized all model parameters with a learning rate of 1e-4 and a weight decay rate of 5e-4 over 300 epochs, using a dropout rate and edge dropout rate of 0.3. Early stopping with a patience of 100 epochs was employed to prevent overfitting.

During feature alignment, the dimensions of imaging and non-imaging features are downsampled or upsampled to 500 dimensions, respectively. The GTUNets encoder depths for imaging and non-imaging features are set to 2 and 3, respectively, with a graph pooling ratio of 0.8. For the ABIDE dataset, the hyperparameters $\lambda $, $\mu$, and $\eta$ in \Cref{eq:18,eq:19} are set to 1, 1e-4, and 1e-2, respectively. For the ADHD-200 dataset, these hyperparameters are set to 1, 1e-1, and 1e-2, respectively.

\subsection{Competitive Methods}
\label{subsec:4.3}

\subsubsection{Performance Evaluation}
\label{subsubsec:4.3.1}
We evaluate the performance of the proposed method on the ABIDE (HC vs. ASD) and ADHD-200 (HC vs. ADHD) datasets using 10-fold stratified cross-validation, while the training, validation and test sets are split in a ratio of 8:1:1, respectively. Namely, in each fold, the dataset is rotated so that each subset serves as the test set once. The performance evaluation metrics, i.e., accuracy (ACC), sensitivity (SEN), specificity (SPE) and area under the ROC curve (AUC), are reported as the average of the 10-fold results on test datasets. 

\subsubsection{Baselines}
\label{subsubsec:4.3.2}
We compared the proposed MM-GTUNets with a few traditional machine learning methods and several SOTA methods in disease prediction tasks. The traditional machine learning methods include support vector machine (SVM) and MLP, where the FC matrix was extracted as a one-dimensional vector and used as input after dimensionality reduction using RFE. The competing SOTA methods include Brain-GNN\cite{li2021braingnn}, DGCN\cite{zhao2022dynamic}, AL-NEGAT\cite{chen2024adversarial}, A-GCL\cite{zhang2023agcl}, Pop-GCN\cite{parisot2018disease}, GATE\cite{peng2023gate}, EV-GCN\cite{huang2022disease}, and MMGL\cite{zheng2022multimodal}.

\begin{table}[ht]
\renewcommand{\arraystretch}{1.2}
\centering
\scriptsize % 调整字体大小为脚注级别
\caption{Performance comparison of different methods on ABIDE and ADHD-200 datasets. "MM" indicates multi-modal data usage: "$\times$" for single-modal, "$\checkmark$" for multi-modal. (bold: optimal, \underline{underline}: suboptimal)}
\resizebox{\columnwidth}{!}{
\begin{tabular}{p{2.6cm}|c|cccc|cccc}
    \Xhline{1pt}
    \multirow{3}{*}{\textbf{Method}} & \multirow{3}{*}{\textbf{MM}} & \multicolumn{4}{c|}{\textbf{ABIDE}} & \multicolumn{4}{c}{\textbf{ADHD-200}} \\ \cline{3-10}
        & & \multicolumn{4}{c|}{HC vs. ASD} & \multicolumn{4}{c}{HC vs. ADHD} \\ \cline{3-10}
        & & {\textbf{ACC (\%)}} & {\textbf{SEN (\%)}} & {\textbf{SPE (\%)}} & {\textbf{AUC (\%)}} & {\textbf{ACC (\%)}} & {\textbf{SEN (\%)}} & {\textbf{SPE (\%)}} & {\textbf{AUC (\%)}} 
         \\ \hline
    (T)SVM & $\times$ & 66.02 (0.35) & 65.34 (0.26) & 78.83 (0.43) & 64.97 (0.37) & 66.48 (0.16) & 64.56 (2.16) & 22.04 (0.62) & 57.59 (0.21) \\ 
    (T)MLP & $\times$ & 72.69 (0.84) & 72.02 (2.84) & 73.45 (0.61) & 77.69 (1.02) & 75.28 (0.33) & 62.81 (3.59) & 82.64 (0.47) & 82.03 (0.40) \\ 
    (B)Brain-GNN\cite{li2021braingnn} & $\times$ & 69.76 (3.80) & 67.47 (3.10) & 73.28 (3.26) & 72.50 (3.10) & 65.26 (3.60) & 68.59 (3.30) & 63.05 (4.00) & 66.02 (5.50) \\ 
    (B)DGCN\cite{zhao2022dynamic} & $\times$ & 71.83 (2.90) & 70.90 (2.59) & 71.80 (3.10) & 72.45 (2.98) & 68.81 (3.69) & 69.08 (4.07) & 69.53 (4.71) & 70.05 (4.57) \\ 
    (B)AL-NEGAT\cite{chen2024adversarial} & $\checkmark$ & 73.17 (3.32) & 78.18 (2.66) & 73.28 (3.14) & 75.02 (2.56) & 69.15 (3.82) & 69.28 (4.32) & 72.26 (3.92) & 68.25 (3.69) \\ 
    (B)A-GCL\cite{zhang2023agcl} & $\times$ & 79.04 (2.40) & 81.42 (3.03) & \underline{80.95 (3.19)} & \underline{82.86 (2.91)} & 80.11 (4.30) & \textbf{82.04 (4.58)} & 80.08 (4.10) & 78.78 (4.39) \\ 
    (P)Pop-GCN\cite{parisot2018disease} & $\checkmark$ & 68.43 (0.86) & 78.32 (0.89) & 57.51 (5.99) & 73.90 (3.08) & 75.45 (0.32) & 50.99 (1.96) & \underline{90.11 (0.22)} & 81.72 (1.40) \\ 
    (P)GATE\cite{peng2023gate} & $\times$ & 74.65 (2.50) & 75.59 (2.43) & 76.87 (2.27) & 76.87 (2.27) & 72.20 (0.23) & 77.27 (5.20) & 72.40 (3.78) & 74.61 (3.30) \\ 
    (P)EV-GCN\cite{huang2022disease} & $\checkmark$ & \underline{80.95 (0.27)} & \underline{83.74 (0.61)} & 77.69 (0.18) & 82.37 (0.29) & \underline{80.95 (0.96)} & 60.33 (7.64) & \textbf{93.11 (0.13)} & \underline{87.15 (1.15)} \\ 
    (P)MMGL\cite{zheng2022multimodal} & $\checkmark$ & 80.34 (0.18) & 80.25 (0.21) & 77.16 (3.15) & 79.83 (1.06) & 77.59 (3.64) & 76.21 (2.98) & 74.57 (1.25) & 78.47 (2.29) \\ 
    \textbf{MM-GTUNets (Ours)} & $\checkmark$ & \textbf{82.92 (0.54)} & \textbf{84.22 (0.81)} & \textbf{81.43 (0.64)} & \textbf{88.21 (0.61)} & \textbf{82.68 (0.60)} &  \underline{77.58 (2.11)}  & 85.76 (0.46) & \textbf{90.71 (0.72)} \\ 
    \Xhline{1pt}
\end{tabular}
}
\label{tab:2}
\end{table}

\subsubsection{Qualitative Results}
\label{subsubsec:4.3.3}
\Cref{tab:2} reports the quantitative performance of MM-GTUNets, where the prefixes “(T)”, “(B)”, and “(P)” denote traditional machine learning methods, brain-graph-based methods, and population-graph-based methods, respectively. Based on \Cref{tab:2}, we can see that: \textbf{(i)} Compared to brain-graph-based methods, population-graph-based methods exhibit more stable performance, with smaller standard deviations in performance metrics. This stability may result from brain-graph methods focusing on local brain region features for each subject, while population-graph methods emphasize global association features within the subject population \cite{bessadok2023graph}. \textbf{(ii)} Most multi-modal methods outperform single-modal ones by integrating multiple data sources. Each modality provides unique information, enabling the model to capture a broader range of details. \textbf{(iii)} MM-GTUNets demonstrated outstanding performance on both datasets. For the ABIDE dataset, MM-GTUNets outperformed all comparison baselines across all metrics. For the ADHD-200 dataset, MM-GTUNets achieved the best results in all metrics except sensitivity and specificity.

\subsubsection{Visualization of Modality-Joint Representation}
\label{subsubsec:4.3.4}
\begin{figure}[!t]
\centering
    \subfigbottomskip=2pt %两行子图之间的行间距
    \subfigcapskip=-5pt %设置子图与子标题之间的距离
    \subfigure[ABIDE]{\label{subfig:4a}
    		\includegraphics[width=0.4\linewidth]{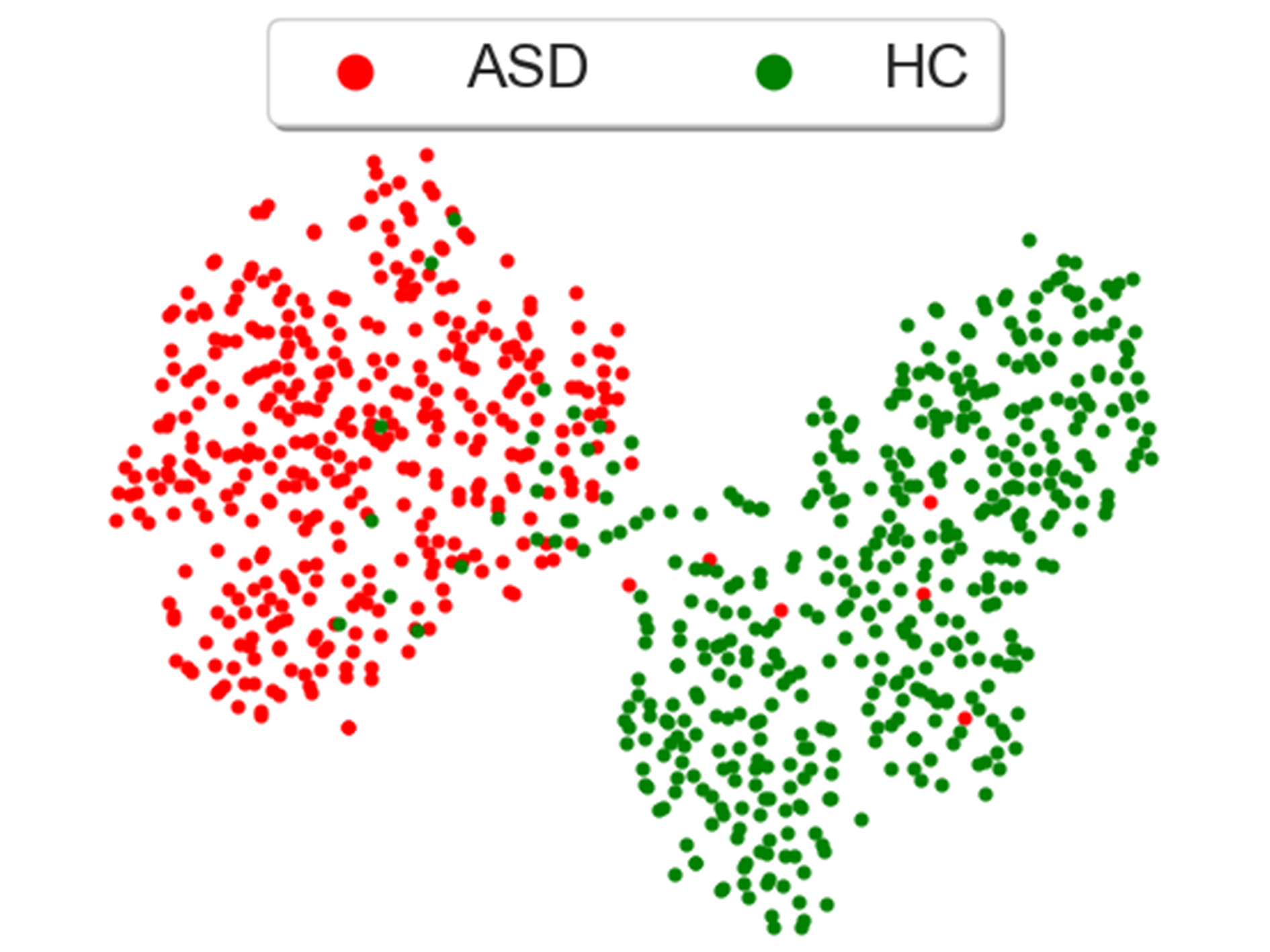}}
    \subfigure[ADHD-200]{\label{subfig:4b}
    		\includegraphics[width=0.4 \linewidth]{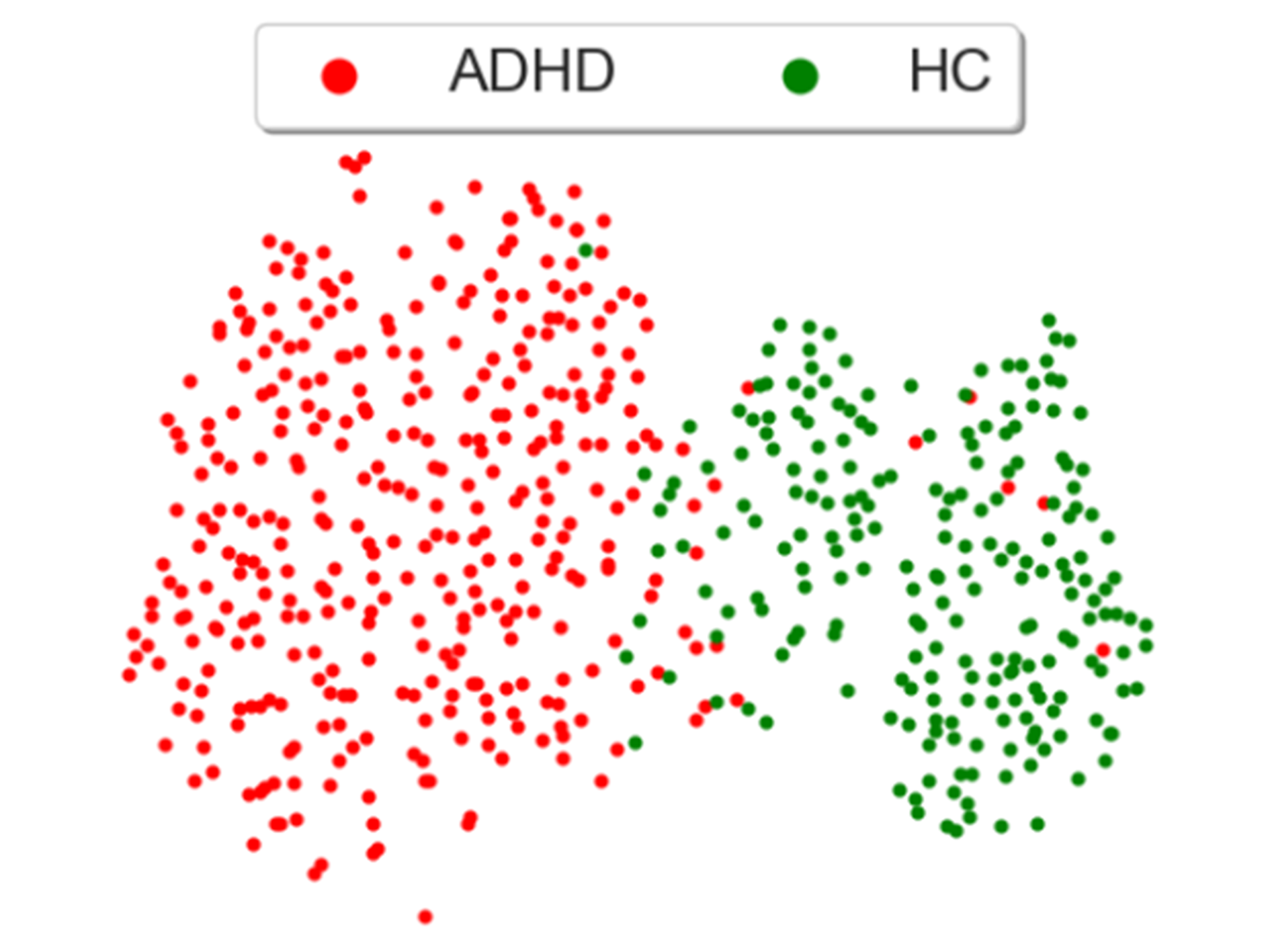}}
\caption{Visualization of the joint representation of modalities.}
\label{fig:4}
\end{figure}

To assess the capability of MM-GTUNets in learning cross-modality interactions, we utilized t-SNE \cite{maaten2008visualizing} to visualize the modality-joint representation $\mathbf{Z}$ in a two-dimensional space for the ABIDE and ADHD-200 datasets. As shown in \Cref{fig:4}, $\mathbf{Z}$ forms two distinct clusters corresponding to the categories, indicating that the multi-modal features learned by MM-GTUNets exhibit significant discriminative power, with low intra-class and high inter-class dispersion. 

\subsection{Ablation Studies}
\label{subsec:4.4}

\subsubsection{Effectiveness of Non-imaging Feature Reconstructor}
\label{subsubsec:4.4.1}
To assess the effectiveness of the proposed non-imaging feature reconstructor in MMRL (\Cref{subsec:3.2}), we compared the performance of different reconstruction methods (VAE, MLP, and autoencoder (AE)) across two datasets. As shown in \Cref{tab:3}, the MMRL module performs poorly without the non-imaging feature reconstructor, while the VAE significantly outperforms the other two methods across both datasets.

\begin{table}
\centering
\renewcommand{\arraystretch}{1.25}
\caption{Ablation study of non-imaging feature reconstructor in MM-GTUNets.(bold:optimal, \customunderline{underline}:suboptimal)}
\begin{tabular}{l|cc|cc}
    \Xhline{1pt}
    \multirow{2}{*}{\textbf{Reconstructor}} & \multicolumn{2}{c|}{\textbf{ABIDE}} & \multicolumn{2}{c}{\textbf{ADHD-200}} \\ \cline{2-5}
     & \textbf{ACC (\%)} & \textbf{AUC (\%)} & \textbf{ACC (\%)} & \textbf{AUC (\%)} \\ 
    \hline
    MRRL w/o VAE & 67.06 (3.27) & 69.99 (4.26) & 76.47 (1.93) & 83.72 (3.07) \\
    MRRL + MLP & \customunderline{80.86 (0.84)} & \customunderline{87.26 (1.20)} & \customunderline{79.04 (0.23)} & \customunderline{89.95 (0.23)} \\
    MRRL + AE & 76.02 (1.59) & 82.42 (2.08) & 77.86 (0.37) & 89.85 (0.36) \\
    MRRL w/ VAE & \textbf{82.92 (0.54)} & \textbf{88.21 (0.61)} & \textbf{82.68 (0.60)} & \textbf{90.71 (0.72)} \\
    \Xhline{1pt}
\end{tabular}
\label{tab:3}
\end{table}

\subsubsection{Effectiveness of Encoder Architecture}
\label{subsubsec:4.4.2}
To investigate the impact of encoder architectures on MM-GTUNets performance, we evaluated the GT layer using stacking architecture, residual architecture, cascade architecture and Graph Unets architecture. As shown in \Cref{tab:4}, among these four architectures, the GTUNet with the Graph Unets achieved the best performance, confirming the effectiveness of the GTUNet architecture introduced in \Cref{subsec:3.3}.

\begin{table}
\centering
\renewcommand{\arraystretch}{1.25}
\caption{Ablation study of the encoder architecture in MM-GTUNets.(bold:optimal, \customunderline{underline}:suboptimal)}
\begin{tabular}{l|cc|cc}
    \Xhline{1pt}
    \multirow{2}{*}{\textbf{Architecture}} & \multicolumn{2}{c|}{\textbf{ABIDE}} & \multicolumn{2}{c}{\textbf{ADHD-200}} \\ \cline{2-5}
     & \textbf{ACC (\%)} & \textbf{AUC (\%)} & \textbf{ACC (\%)} & \textbf{AUC (\%)} \\
    \hline
    Stacking & 77.06 (0.47) & 84.03 (0.71) & 76.14 (0.61) & 86.21 (0.81) \\
    Residual & 78.78 (0.61) & 85.56 (0.79) & 76.33 (0.69) & 88.26 (0.58) \\
    Cascade & \customunderline{79.70 (0.47)} & \customunderline{86.78 (0.57)} & \customunderline{80.97 (0.54)} & \customunderline{89.37 (0.61)} \\
    GTUNet & \textbf{82.92 (0.54)} & \textbf{88.21 (0.61)} & \textbf{82.68 (0.60)} & \textbf{90.71 (0.72)} \\
    \Xhline{1pt}
\end{tabular}
\label{tab:4}
\end{table}

\subsubsection{Impact of Embedding Dimensions and Pooling Ratios}
\label{subsubsec:4.4.3}
\begin{figure}[!t]
\centering
    \subfigbottomskip=0pt %两行子图之间的行间距
    \subfigcapskip=-5pt %设置子图与子标题之间的距离
    \subfigure[Impact of embedding dimensions]{\label{subfig:5a}
    		\includegraphics[width=0.45\linewidth]{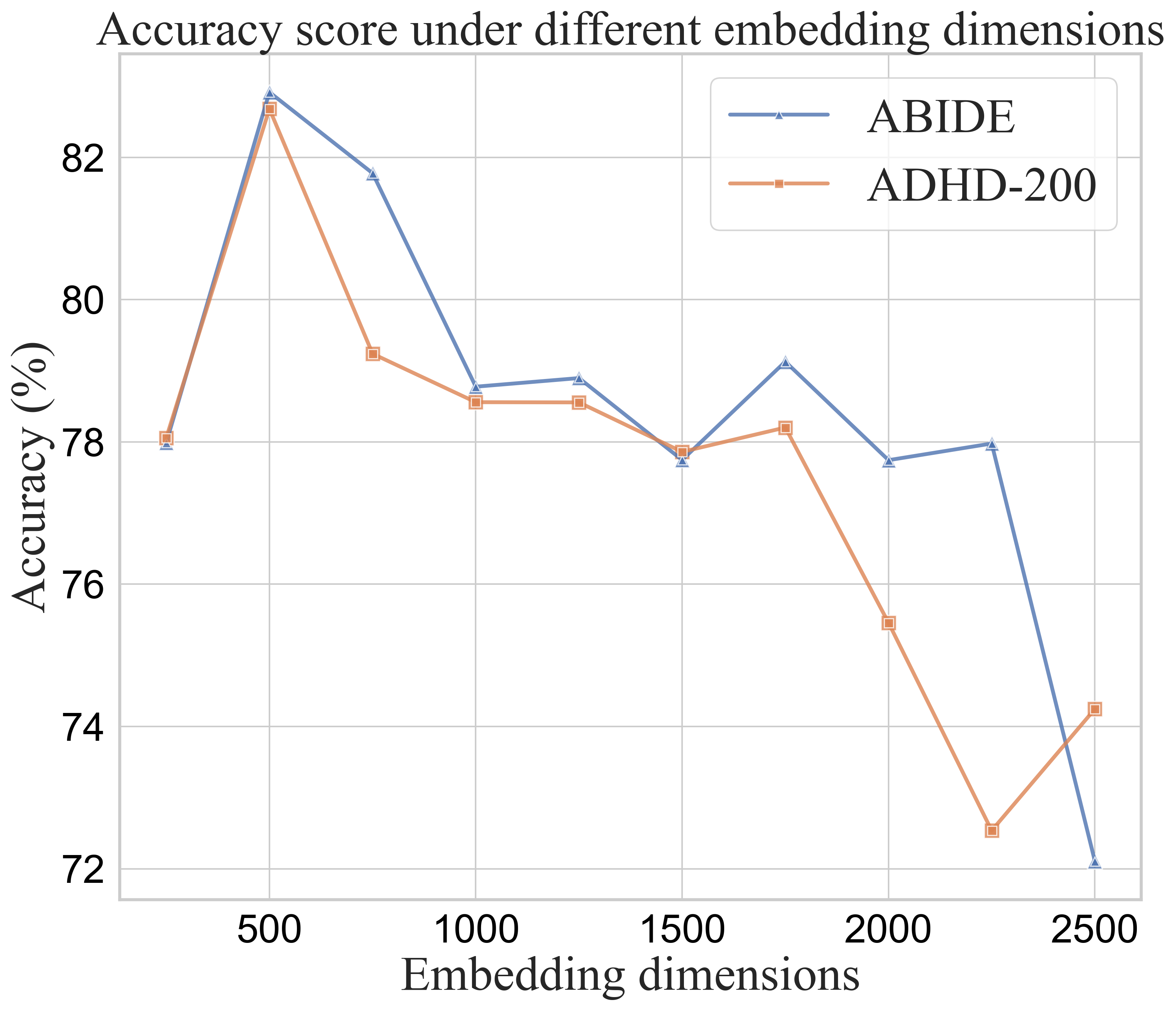}}
    \subfigure[Impact of pooling ratios]{\label{subfig:5b}
    		\includegraphics[width=0.45\linewidth]{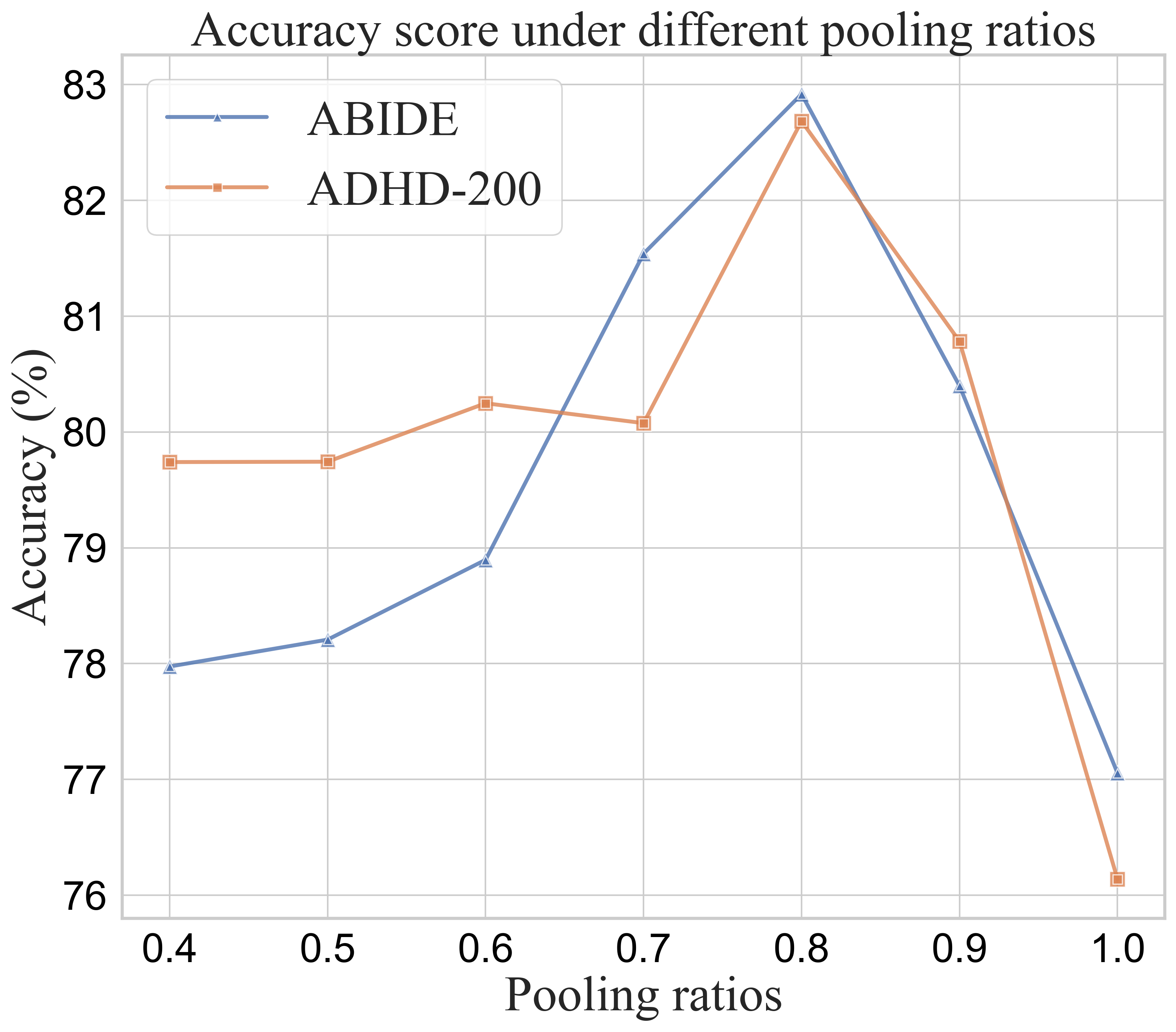}}
\caption{Accuracy of MM-GTUNets with different embedding dimensions and different pooling ratios.}
\label{fig:5}
\end{figure}

To analyze the impact of embedding dimensions of multi-modal features on MM-GTUNets' performance, we tested a range of 250 to 2500 with step length equal to 250. As shown in \Cref{subfig:5a}, different from the findings in \cite{parisot2018disease,huang2022disease,peng2023gate}, on both datasets, the performance of MM-GTUNets peaked at an embedding dimension of 500 and then gradually declined. This decline may result from excessively high embedding dimensions causing redundant multi-modal feature representations, which degrades the performance.

To reveal the impact of the graph pooling ratios of the encoder GTUNet, we tested a range of 0.4 to 1.0 with step length equal to  0.1. Similarly, as shown in \Cref{subfig:5b}, in both datasets, the performance of MM-GTUNets peaked at a pooling ratio of 0.8 and then gradually declined, reaching its lowest point at a pooling ratio of 1.0. Low pooling ratios may remove important node features and edges, causing the loss of critical information, while high pooling ratios may neglect key features and significantly reduce the performance.

\subsubsection{Effectiveness of MRRL}
\label{subsubsec:4.4.4}
\begin{figure}[!t]
\centering
    \subfigbottomskip=0pt %两行子图之间的行间距
    \subfigcapskip=-5pt %设置子图与子标题之间的距离
    \subfigure[ABIDE]{\label{subfig:6a}
    		\includegraphics[width=0.45\linewidth]{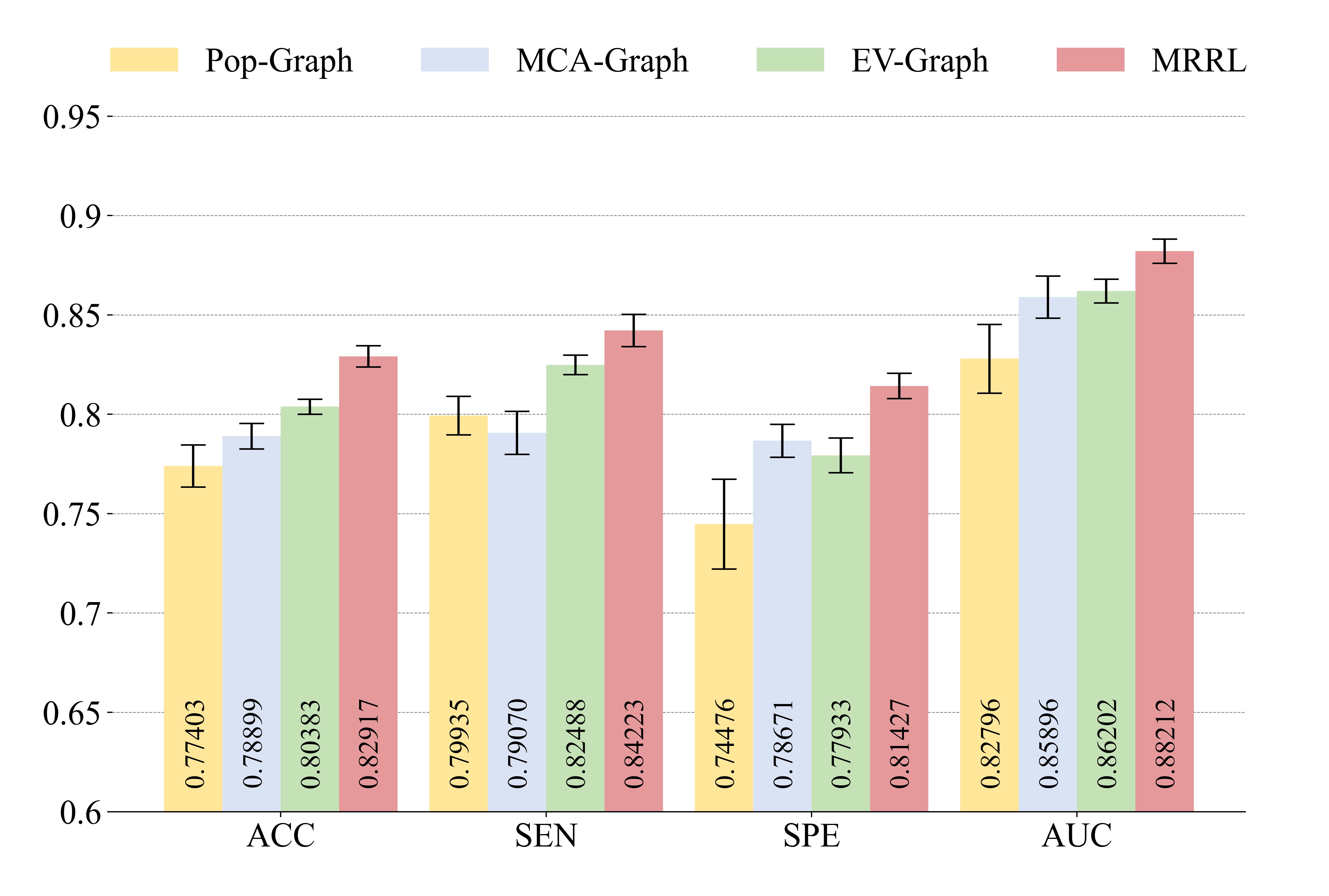}}
    \subfigure[ADHD-200]{\label{subfig:6b}
    		\includegraphics[width=0.45\linewidth]{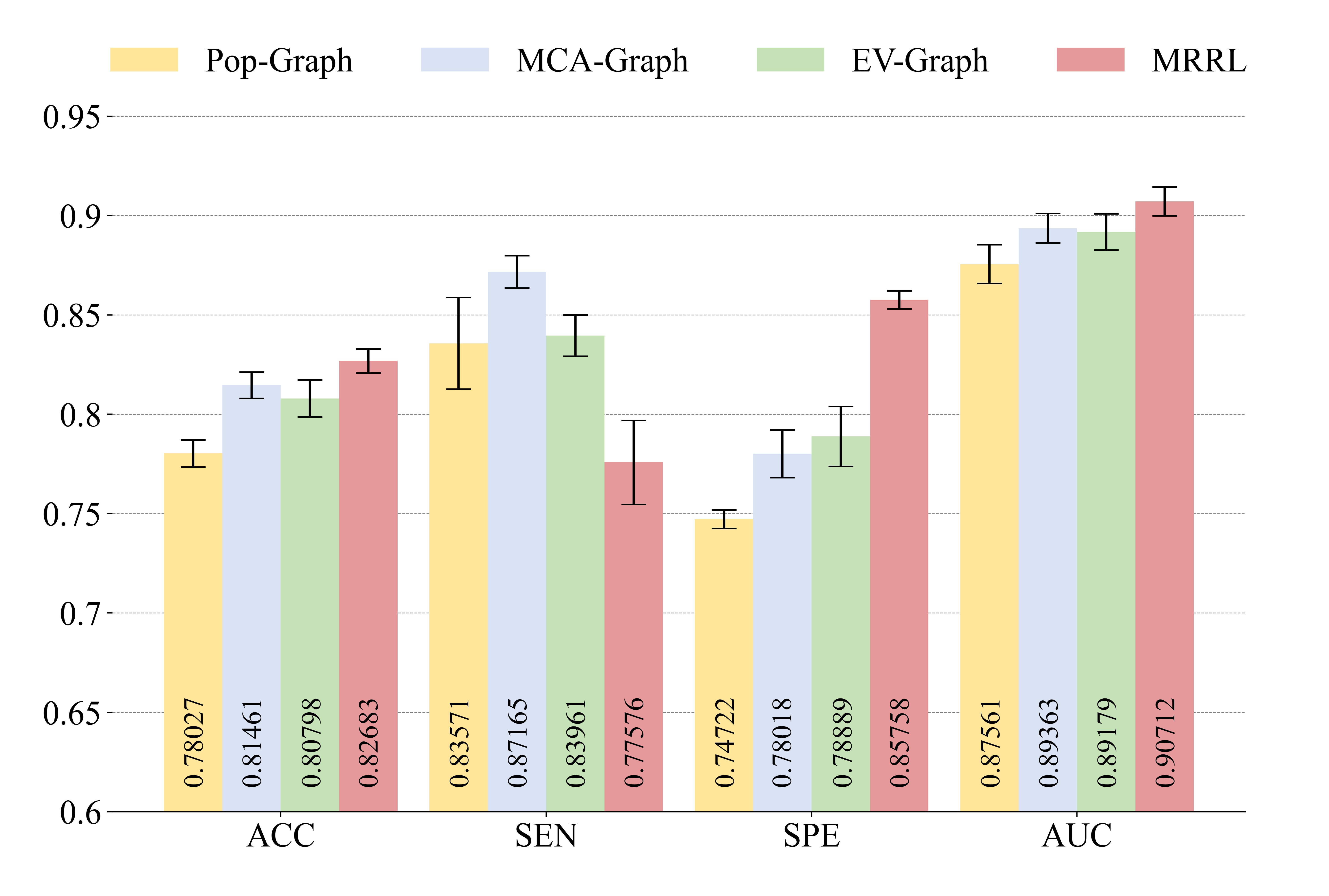}}
\caption{Ablation study of graph construction methods in MM-GTUNets.}
\label{fig:6}
\end{figure}

To evaluate the effectiveness of the adaptive reward-based population graph construction, we compared our MRRL method with other population graph construction methods, i.e., Pop-Graph (static)\cite{parisot2018disease}, EV-Graph (adaptive)\cite{huang2022disease}, and MCA-Graph (adaptive)\cite{song2023multicenter}. In this ablation study, we replaced the AMRS in the MRRL module with the non-imaging affinity calculation methods from Pop-Graph, EV-Graph, and MCA-Graph, respectively. As shown in \cref{fig:6}, the manually constructed Pop-Graph performs the worst among all methods. Compared to adaptive graph construction baselines like EV-Graph and MCA-Graph, our MRRL shows overall better performance.

\subsubsection{Contribution of Non-Imaging Data}
\label{subsubsec:4.4.5}
Non-imaging data has proven essential for predicting brain disorders based on MMGDL \cite{huang2022disease, pellegrini2023unsupervised}. As shown in \Cref{tab:5}, our MM-GTUNets achieves optimal performance when both imaging and non-imaging data are incorporated as inputs. Excluding non-imaging data results in a noticeable decline in MM-GTUNets’ performance, while excluding imaging data entirely eliminates its classification capability. These results indicate that non-imaging data effectively complements imaging data, enhancing MM-GTUNets’ overall performance, with imaging data exerting a dominant influence on the framework’s effectiveness.

\begin{table}[ht]
\centering
\renewcommand{\arraystretch}{1.25}
\caption{Impact of input data on MM-GTUNets' performance. (bold:optimal, \customunderline{underline}:suboptimal)}
\begin{tabular}{l|cc|cc}
    \Xhline{1pt}
    \multirow{2}{*}{\textbf{Input Data}} & \multicolumn{2}{c|}{\textbf{ABIDE}} & \multicolumn{2}{c}{\textbf{ADHD-200}} \\ \cline{2-5}
     & \textbf{ACC (\%)} & \textbf{AUC (\%)} & \textbf{ACC (\%)} & \textbf{AUC (\%)} \\
    \hline
    Non-Imaging & 52.00 (0.24) & 50.28 (0.38) & 58.07 (0.23) & 53.74 (0.40) \\
    Imaging & \customunderline{76.49 (0.80)} & \customunderline{85.22 (0.72)} & \customunderline{79.58 (0.66)} & \customunderline{87.66 (0.88)} \\
    Imaging + Non-Imaging & \textbf{82.92 (0.54)} & \textbf{88.21 (0.61)} & \textbf{82.68 (0.60)} & \textbf{90.71 (0.72)} \\
    \Xhline{1pt}
\end{tabular}
\label{tab:5}
\end{table}

\subsection{Scalability and Hardware Requirements}
\label{subsec:4.5}
\begin{figure}[!t]
\centering
    \subfigbottomskip=0pt %两行子图之间的行间距
    \subfigcapskip=-5pt %设置子图与子标题之间的距离
    \subfigure[ABIDE]{\label{subfig:7a}
    		\includegraphics[width=0.45\linewidth]{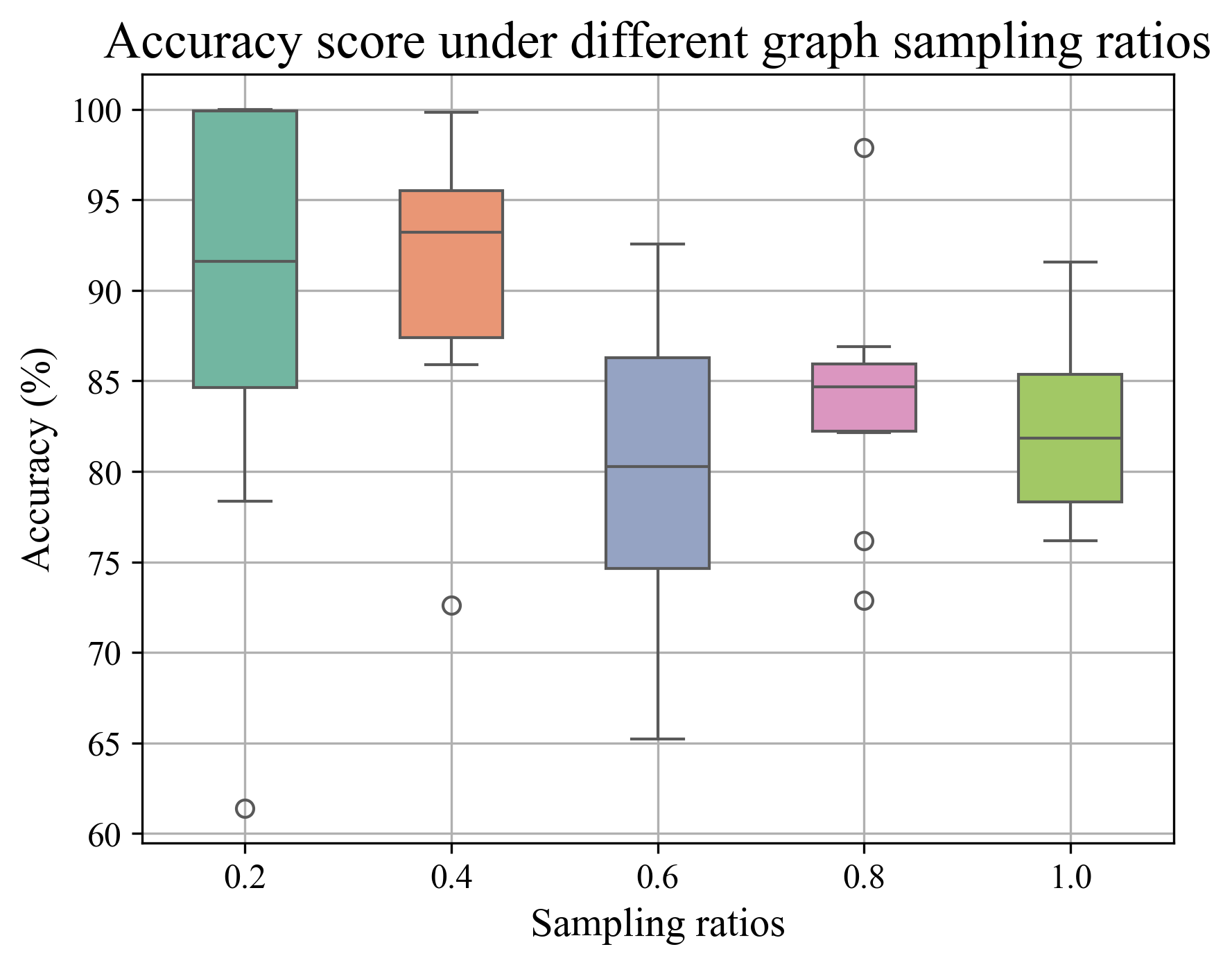}}
    \subfigure[ADHD-200]{\label{subfig:7b}
    		\includegraphics[width=0.45\linewidth]{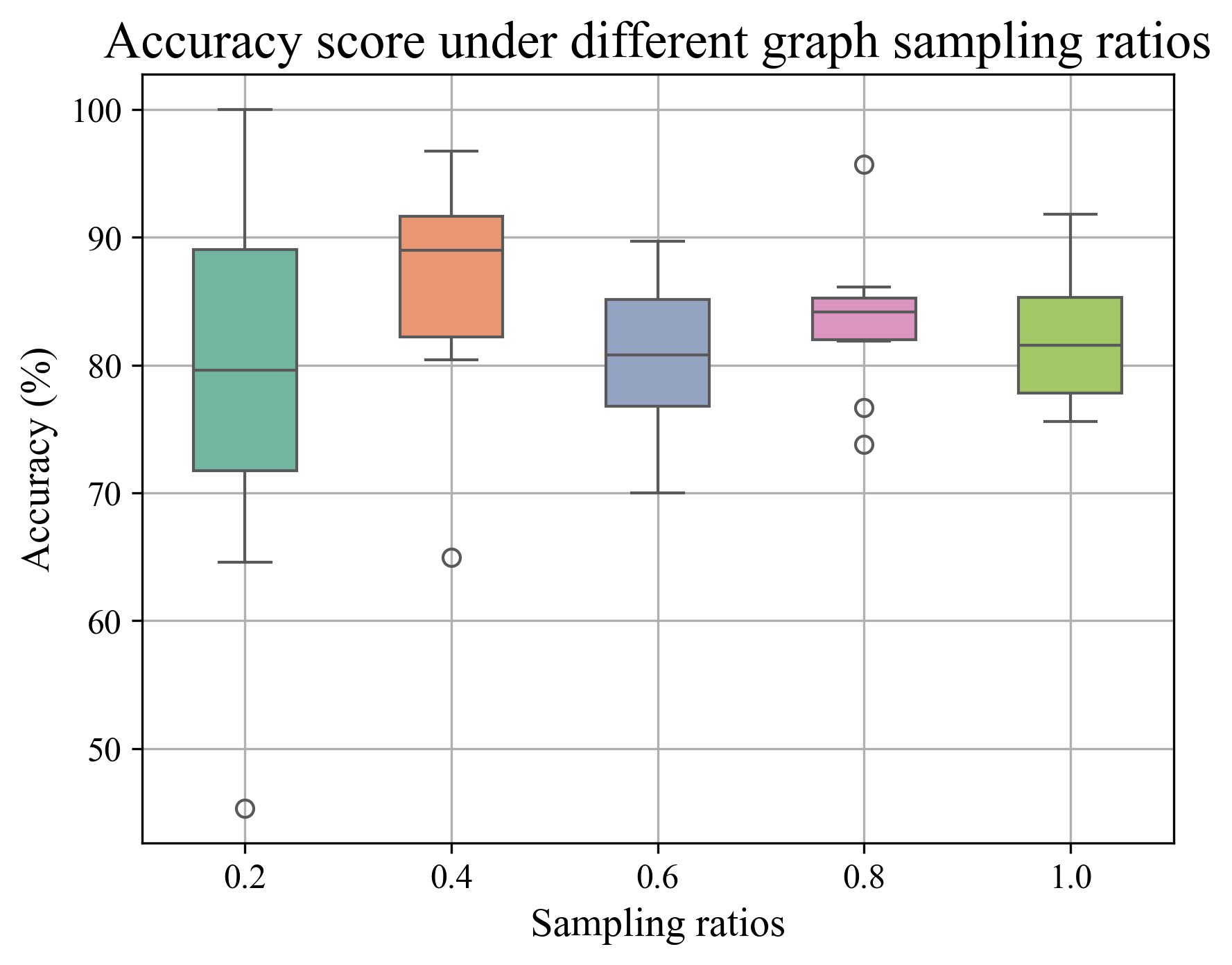}}
\caption{Accuracy of MM-GTUNets at different graph sampling ratios on ABIDE and ADHD-200.}
\label{fig:7}
\end{figure}

To evaluate the scalability of our proposed MM-GTUNets, we performed multiple rounds of stratified sampling on the experimental datasets, with sampling ratios ranging from 0.2 to 1.0 in increment step of 0.2. We then compared MM-GTUNets’ performance across varying graph sizes. As shown in \Cref{fig:7}, the results clearly indicate that as the graph sampling ratio increases (i.e., the graph size grows), MM-GTUNets' performance gradually converges and eventually stabilizes. Additionally, we report the floating-point operations per second (FLOPs), maximum GPU memory usage, and average training time of MM-GTUNets under different graph sizes.  As shown in \Cref{tab:6}, the hardware requirements of MM-GTUNets grow with the graph size.

\begin{table}[ht]
\centering
\renewcommand{\arraystretch}{1.5}
\caption{Hardware requirements of MM-GTUNets at different graph sampling ratios.}
\resizebox{\linewidth}{!}{
\begin{tabular}{c|cccc|cccc}
    \Xhline{1pt}
    \multirow{2}{*}{\textbf{Sampling Ratio}} & \multicolumn{4}{c|}{\textbf{ABIDE}} & \multicolumn{4}{c}{\textbf{ADHD-200}} \\ \cline{2-9}
     & \textbf{Memory (MB)} & \textbf{FLOPs (G)} & \textbf{Time (s)} &\textbf{Time\textsuperscript{*} (s)} &\textbf{Memory (MB)} & \textbf{FLOPs (G)} & \textbf{Time (s)} & \textbf{Time\textsuperscript{*} (s)} \\
    \hline
    0.2 & 1126 & 0.07 & 23.10 & 19.77 & 787 & 0.05 & 22.02 & 16.37  \\
    0.4 & 2221 & 0.14 & 28.99 & 24.59 & 1259 & 0.09 & 22.82 & 17.62 \\
    0.6 & 4407 & 0.20 & 44.58 & 37.67 & 2295 & 0.14 & 27.92 & 22.62 \\
    0.8 & 7217 & 0.27 & 67.16 & 60.28 & 3539 & 0.18 & 37.92 & 29.81 \\
    1.0 & 11335 & 0.34 & 94.99 & 89.11 & 5303 & 0.23 & 51.10 & 47.53 \\
    \Xhline{1pt}
\multicolumn{9}{l}{Note: \textbf{Time} denotes the training time without early stopping, while \textbf{Time\textsuperscript{*}} denotes the training time with early stopping.}\\
\end{tabular}
}
\label{tab:6}
\end{table}

\section{DISCUSSION}
\label{sec:5}

\subsection{Transfer Learning Capabilities of VAE}
\label{subsec:5.1}
In \Cref{subsec:3.2}, we state that VAE can bridge the modal gap between imaging and non-imaging data and demonstrate its effectiveness in \Cref{subsec:4.4}. However, in practical applications, due to the scarcity of available data sources, the data for pretraining is often limited. Leveraging transfer learning for VAE feature reconstruction presents a promising solution to this issue\cite{hou2022variational, bonheme2023good}. Therefore, we further investigated the effect of transfer learning with VAE on the two datasets used. As shown in \Cref{tab:7}, we evaluated the performance of MM-GTUNets after pretraining the VAE on different datasets, ensuring that the non-imaging data categories used for pre-training were consistent. The results indicate that the VAE pretrained on the ABIDE dataset with a larger scale may exhibit better generalization performance.

\begin{table}[ht]
\centering
\renewcommand{\arraystretch}{1.25}
\caption{Performance of VAE with Pre-Training on Various Datasets (bold:optimal)}
\begin{tabular}{l|cc|cc}
    \Xhline{1pt}
    \multirow{2}{*}{\textbf{Pretrained Datasets}} & \multicolumn{2}{c|}{\textbf{ABIDE}} & \multicolumn{2}{c}{\textbf{ADHD-200}} \\ \cline{2-5}
     & \textbf{ACC (\%)} & \textbf{AUC (\%)} & \textbf{ACC (\%)} & \textbf{AUC (\%)} \\
    \hline
    ABIDE & \textbf{82.92 (0.54)} & \textbf{88.21 (0.61)} & \textbf{83.02 (0.53)} & \textbf{91.11 (0.81)} \\
    ADHD-200 & 81.18 (0.27) & 87.71 (0.46) & 82.68 (0.60) & 90.71 (0.72) \\
    \Xhline{1pt}
\end{tabular}
\label{tab:7}
\end{table}

\subsection{Impact of Imaging Data Pre-Processing Techniques}
\label{subsec:5.2}
Different imaging data preprocessing techniques can sometimes lead to varying conclusions from the same neuroimaging datasets\cite{luppi2024systematic}. To examine the impact of imaging data preprocessing techniques on MM-GTUNets' performance, we assessed the framework using the ABIDE dataset under various preprocessing pipelines. As is shown in \Cref{tab:8}, the performance of MM-GTUNets on the ABIDE dataset is not sensitive to data preprocessing techniques, demonstrating the robustness of our proposed framework.

\begin{table}[ht]
\centering
\renewcommand{\arraystretch}{1.25}
\caption{Performance of MM-GTUNets under different preprocessing pipelines on  ABIDE dataset. (bold:optimal, \customunderline{underline}:suboptimal)}
\begin{tabular}{c|cccc}
    \Xhline{1pt}
     \textbf{Pipeline} & \textbf{ACC (\%)} & \textbf{SEN (\%)} & \textbf{SPE (\%)} & \textbf{AUC (\%)} \\
    \hline
    CCS & 82.22 (0.35) & 82.50 (0.51) & \customunderline{81.93 (0.47)}	& 88.35 (0.44) \\
    DPARSF & \customunderline{82.80 (0.51)} & 83.34 (0.49) &  \textbf{82.18 (0.70)} & \customunderline{88.62 (0.90)} \\
    NIAK & \customunderline{82.80 (0.44)}	& \textbf{84.44 (1.03)} & 80.92 (0.66) & \textbf{89.55 (0.55)} \\
    CPAC & \textbf{82.92 (0.54)} & \customunderline{84.22 (0.81)} & 81.43 (0.64) & 88.21 (0.61) \\

    \Xhline{1pt}
\end{tabular}
\label{tab:8}
\end{table}

\subsection{Interpretability Analysis}
\label{subsec:5.3}
\begin{figure*}[!t]
\centering
    \subfigure[ABIDE]{\label{subfig:8a}
    		\includegraphics[width=0.4\linewidth]{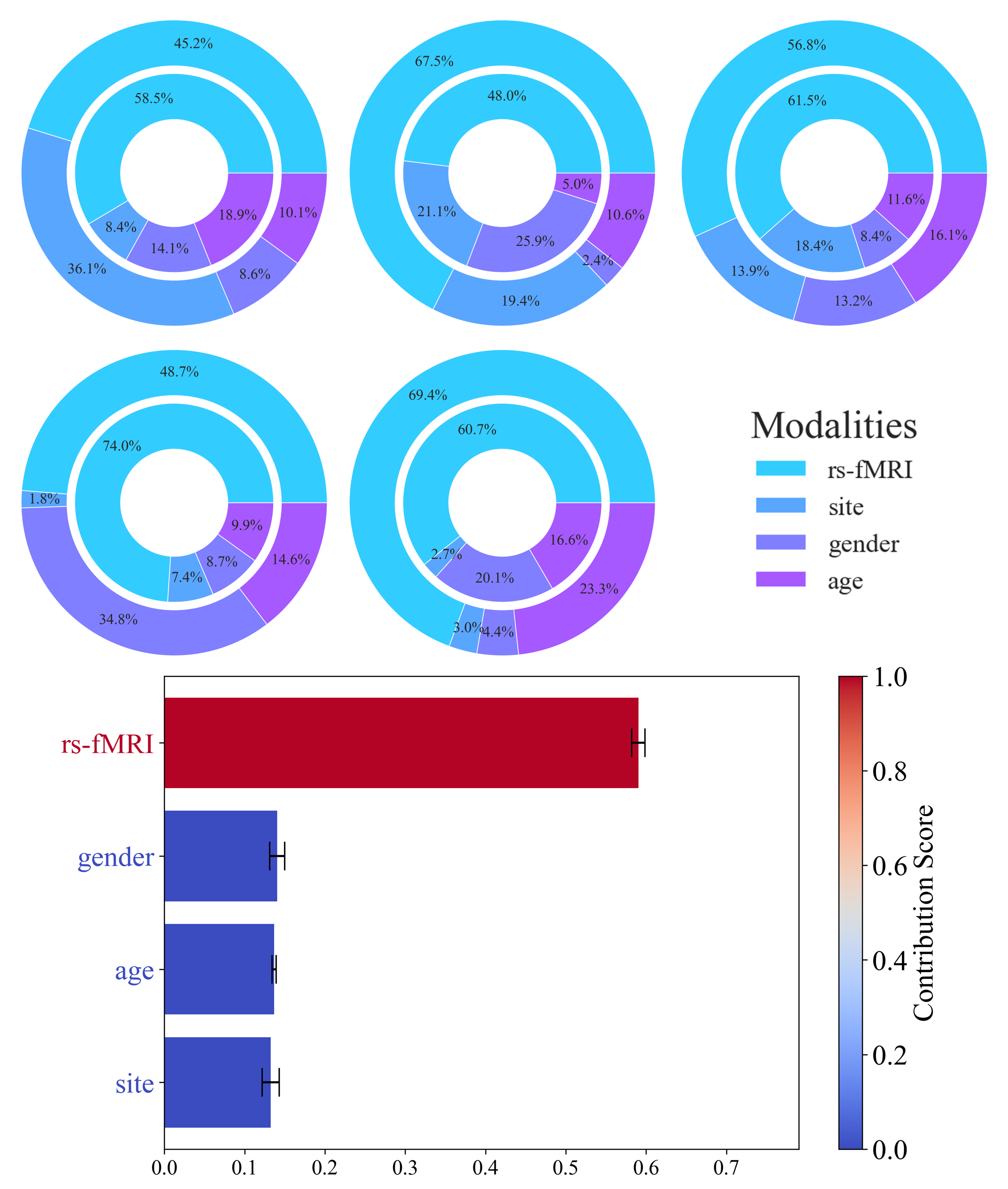}}
    \hspace{0.05\linewidth} % 调整此值来设置子图间的距离
    \subfigure[ADHD-200]{\label{subfig:8b}
    		\includegraphics[width=0.4\linewidth]{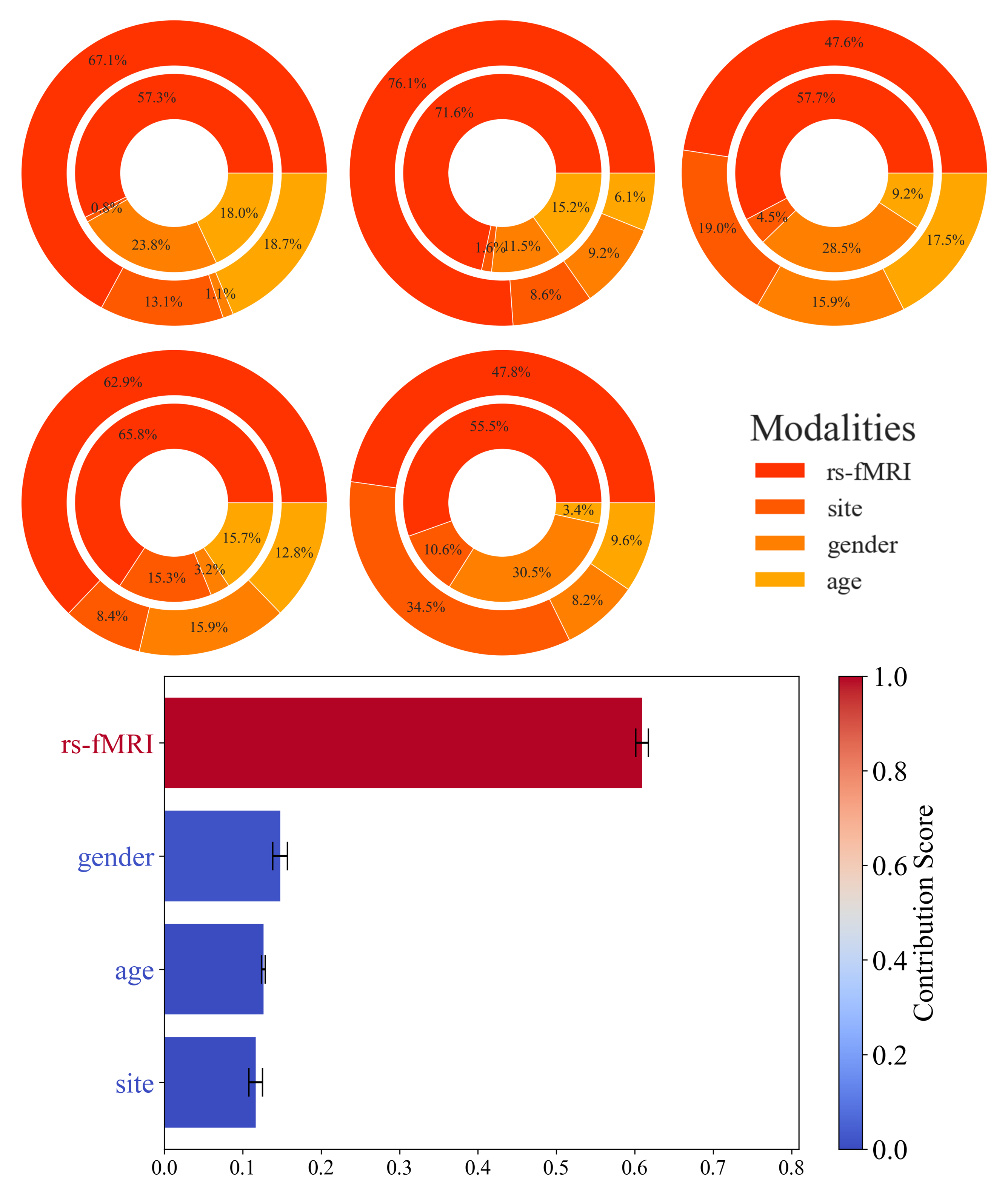}}
\caption{The contribution weight of each modality for classification tasks in 10-fold stratified cross-validation on ABIDE and ADHD-200.}
\label{fig:8}
\end{figure*}

To quantify each modality's contribution in prediction tasks, we calculated the contribution weights for each modality in 10-folds, including various non-imaging data. As shown in \Cref{fig:8}, the average contribution weights on two datasets using MM-GTUNets reveal that rs-fMRI data contributes the most to predictions. Among non-imaging data, the contribution scores for gender, age and site are relatively balanced. Among these factors, gender has a greater influence on prediction results, while age demonstrates a more consistent impact.

\begin{figure}[!t]
\centering
\includegraphics[width=0.7\linewidth]{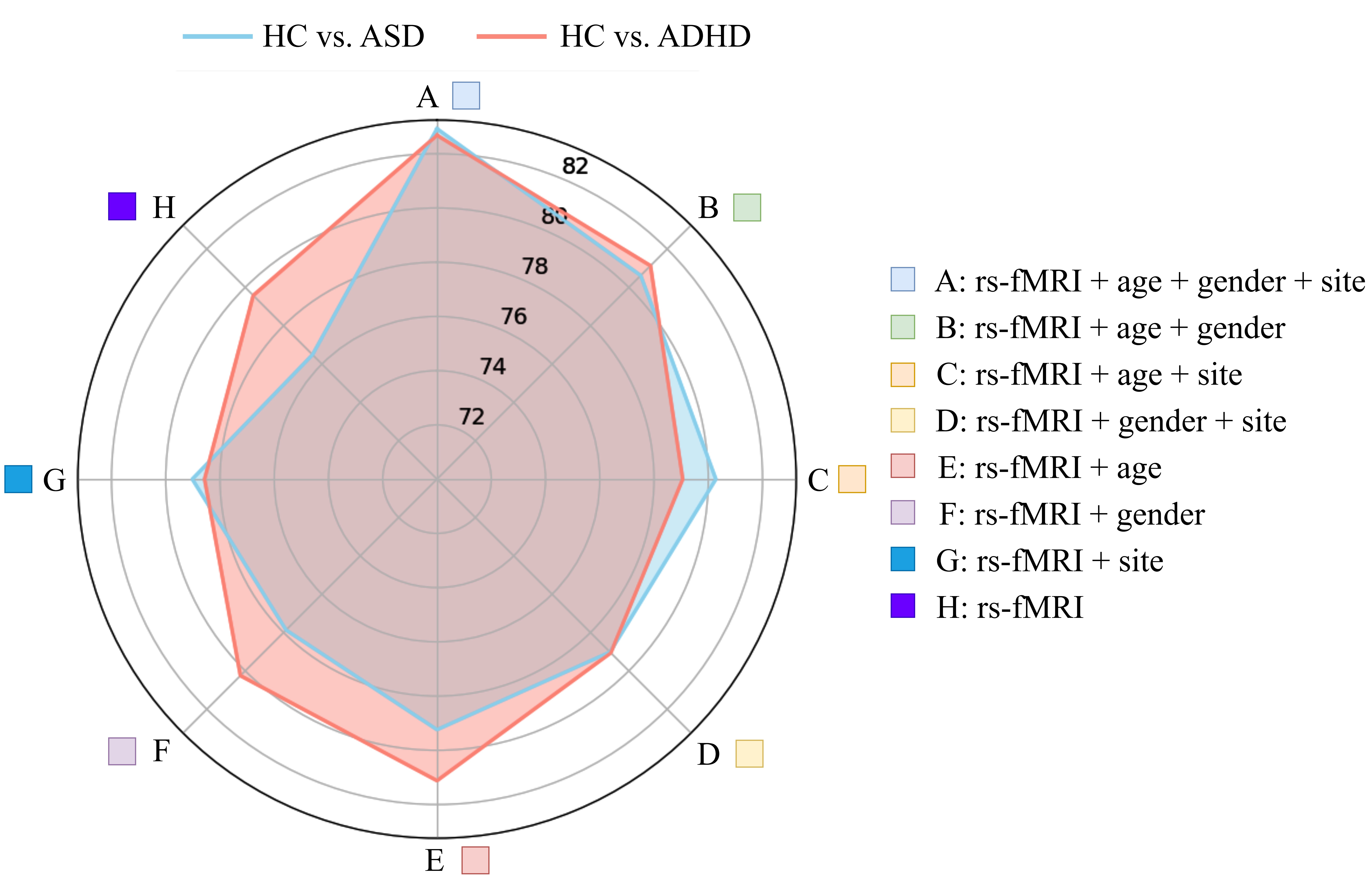}
\caption{The effect of each modality on the classification accuracy on ABIDE and ADHD-200.}
\label{fig:9}
\end{figure}

According to \Cref{fig:9}, the model incorporating rs-fMRI and all non-imaging data types achieves the highest accuracy on both datasets. Among models containing only a single type of non-imaging data, the models containing age perform best. Thus, on ABIDE and ADHD-200, our MM-GTUNets effectively captures complex inter- and intra-modality relationships.

\subsection{Limitations and Future Work}
\label{subsec:5.4}
In this study, certain limitations remain to be addressed in future. 1) Despite MM-GTUNets demonstrating satisfactory performance in binary classification tasks using rs-fMRI data and non-imaging data, studies on multi-classification tasks using more modalities (e.g., TADPOLE dataset\cite{marinescu2018tadpole}) could be further explored. 2) Our approach learns only from samples that have passed quality checks, leaving low-quality multi-modal data (incomplete modalities, noisy and imbalanced) underutilized. Extending our work to low-quality multi-modal learning tasks would be valuable\cite{wang2023distribution, yang2024incomplete, zhang2024multimodal}. 3) Considering that medical personnel often need to make real-time or quick decisions in clinical settings, our framework based on transductive learning may not be suitable for real-time applications due to the need for processing additional test data during prediction. In future work, we will adapt MM-GTUNets to support inductive learning scenarios as described in \cite{zheng2022multimodal, kazi2022differentiable}. 4) Multi-modal fusion aims to capture complex inter- and intra-modal relationships. Building on the insights from Cao et al.\cite{cao2024predictive}, we will devise new approach to improve the ACMGL module, aiming to enhance the quality of fused features and stabilize the weights of modality contributions. 5) To address the hardware challenges arising from the increasing graph size data shown in \Cref{tab:6}, further optimization of MM-GTUNets is required in the future.

\section{CONCLUSION}
\label{sec:6}
In this paper, we propose a unified multi-modal graph deep learning framework named MM-GTUNets for BD prediction based on the graph transformer (GT) to capture complex inter- and intra- modal relationships within large-scale multi-modal data. Importantly, the proposed AMRS helps adaptively learn the population graph and considers the contribution weights of non-imaging features. Furthermore, we propose the GTUNet to extract critical local node embeddings from the global context within the graph, which incorporates the advantages of the Graph Unets architecture and GT operation. Finally, the modality-joint representation is formed by the embedding of different modalities through the feature fusion module. The module can also provide the visualization of the inter- and intra-modal contribution weights learned by the model, which gives our framework the potential to provide modality-interpretable decision support in medical applications.

\bibliographystyle{IEEEtran}
\bibliography{references}

\end{document}